\documentclass[11pt]{article}

\usepackage[preprint]{acl}

\usepackage{times}
\usepackage{latexsym}

\usepackage[T1]{fontenc}

\usepackage[utf8]{inputenc}

\usepackage{microtype}

\usepackage{inconsolata}

\usepackage{graphicx}

\usepackage{amsmath}
\usepackage{amssymb}
\usepackage{booktabs}
\usepackage{multirow}
\usepackage{enumitem}
\AtBeginDocument{\let\origthebibliography\thebibliography
  \renewcommand{\thebibliography}[1]{\origthebibliography{#1}\interlinepenalty=10000\relax}}

\usepackage[normalem]{ulem}

\usepackage{tikz}
\usetikzlibrary{arrows.meta,positioning,calc,fit,backgrounds,shapes.geometric,shapes.arrows}

\newcommand{\Mpriv}{M_{\mathrm{priv}}}          
\newcommand{\Mpub}{M_{\mathrm{pub}}}            
\newcommand{\Medit}{M_{\mathrm{priv}}^{\setminus\mathcal{N}}}       
\newcommand{\LM}{\mathcal{L}_{\mathrm{LM}}}     
\newcommand{\LJEPA}{\mathcal{L}_{\mathrm{JEPA}}} 
\newcommand{\xs}{x^{\mathrm{sur}}}                
\newcommand{\xm}{x^{\mathrm{mask}}}                
\newcommand{\xc}{x^{\mathrm{type}}}                
\newcommand{\Aset}{\mathcal{R}}                 
\newcommand{\Jx}{\mathcal{J}_{x}}               
\newcommand{\htgt}{h^{\mathrm{ref}}_{x}}               
\newcommand{\Chat}{\hat{C}}                     
\newcommand{\method}{\textsc{JEPA-Prune}}       

\title{Prune the Leaky Neurons: Mitigating Private Data Leakage in Fine-Tuned LLMs via Joint-Embedding-Guided Pruning}

\author{Mohamed Shaaban and Mohamed Elmahallawy \\
  School of Engineering and Applied Sciences, Washington State University, Richland, WA 99354, USA \\
  \texttt{\{mohamed.shaaban, mohamed.elmahallawy@wsu.edu\}}
  }

\begin{document}
\maketitle
\begin{abstract}
Fine-tuning large language models (LLMs) on proprietary records enables domain adaptation but introduces two challenges: (i) \emph{privacy leakage} of memorized private data, such as personally identifiable information (PII), and (ii) {\em balancing private data removal} with {\em utility preservation}. Existing defenses, including data sanitization and differential privacy, often degrade performance, while privacy-neuron editing lacks a PII-free reference to distinguish identity-specific neurons from those supporting useful text. We propose \method{}, a \emph{post-hoc neuron pruning framework} guided by a model trained exclusively on PII-free views. Alongside the fine-tuned \emph{private model}, \method{} trains a \emph{public model} on surrogate and masked views of the same records using a joint-embedding predictive architecture (JEPA), capturing domain knowledge without identifiers. A \emph{cross-model representation probe} quantifies deviations between private representations of leaked records and this PII-free reference. Integrated gradients and a scale-invariant specificity score then identify \emph{PII-responsive neurons}, which are pruned without retraining. Experiments show that pruning just 1.6\% of neurons reduces PII inference accuracy by up to 7.4$\times$ and PII extraction recall by up to 13.9$\times$, while keeping unseen-text perplexity within 16\% of that of the unpruned model for most LLMs. In contrast, privacy-neuron editing increases perplexity by at least 26$\times$.
\end{abstract}

\section{Introduction} \label{sec:intro}

Large language models (LLMs) become valuable to organizations when adapted to domains such as legal analysis, financial reporting, and medical question answering \citep{ling2025domain,wu2023bloomberggpt,singhal2025toward}. Since training from scratch is often impractical, organizations fine-tune pretrained LLMs on proprietary records, frequently using lightweight methods such as low-rank adaptation (LoRA) \citep{pratap2025fine,dettmers2023qlora}. These records contain both the domain knowledge that makes models useful and private information about the people they describe.

However, LLMs can memorize training data and reproduce it verbatim \citep{carlini2021extracting,nasr2025scalable}. Fine-tuned models may therefore disclose personally identifiable information (PII), such as names and addresses, to adversaries through ordinary queries \citep{lukas2023analyzing,nakka2025pii}. Such leakage can facilitate targeted phishing \citep{bethany2025lateral} and violate privacy obligations under regulations such as the General Data Protection Regulation (GDPR) \citep{regulation2016regulation}. Because fine-tuning learns identities alongside domain knowledge, organizations face a key challenge: \emph{reducing private data leakage without erasing useful domain knowledge}.

Existing defenses, including data scrubbing \citep{pilan2022text,lukas2023analyzing} and differential privacy (DP) \citep{dwork2006differential,abadi2016deep}, reduce memorization but can compromise utility. Scrubbing masks PII and may remove task-relevant context, while DP injects noise into training updates, potentially degrading accuracy at stronger privacy guarantees \citep{lukas2023analyzing}. Both operate during training, making post-deployment leakage difficult to address without retraining or additional unlearning updates \citep{kuo2025proactive}. Privacy-neuron editing avoids such updates by deactivating neurons associated with PII \citep{wu2023depn}, but analyzes the leaking model alone, \emph{without a PII-free reference to distinguish identity-specific representations from those supporting useful context}. This motivates our central question: \emph{How can an organization identify and remove neurons responsible for private data leakage from a deployed LLM without retraining or sacrificing its domain knowledge?}


We propose \method{}, a post-hoc neuron pruning framework that mitigates private data leakage while preserving domain utility. Alongside the deployed \emph{private model}, fine-tuned on raw records, \method{} trains a \emph{public model} on surrogate and masked views of the same records, where each private data value is replaced with a synthetic substitute or a placeholder. We define this public model as a \emph{PII-free reference}: it captures domain knowledge and contextual information without access to the original PII values. Trained using a joint-embedding predictive architecture (JEPA) objective \citep{huang2026llm}, the public model provides a reference representation against which \method{} compares the private model. Specifically, \method{} reuses the objective's cosine-based formulation as a \emph{cross-model probe} to measure representation differences between the private model's encoding of a leaking record and the corresponding PII-free reference. Integrated gradients attribute these differences to individual neurons, while a scale-invariant specificity score identifies \emph{PII-responsive neurons} whose contributions are stronger on records containing PII than on their placeholder-substituted counterparts. Pruning these neurons requires no retraining or changes to the serving interface. Our key contributions are:
\begin{itemize}[leftmargin=*]
\item \textbf{\em Reference-guided post-hoc privacy protection.} We propose \method{}, which identifies and {\em selectively prunes} neurons associated with {\em memorized private data} by comparing a fine-tuned LLM against a PII-free reference, rather than analyzing the leaking model in isolation.

\item \textbf{\em A PII-free reference on a shared backbone.} We train private and public LoRA adapters on a shared frozen backbone to enable direct representation comparisons: {\sc(i) Private model:} Fine-tuned on raw PII records and deployed after pruning;  {\sc (ii) Public model:} Trained with a JEPA objective on surrogate (syntactic) PII and masked PII views, without access to the original PII, and used exclusively as a reference.

\item \textbf{PII-specific neuron identification.} On leaked records, a cross-model probe measures how far the private model's embedding of a record lies from the public model's embedding of the same record with its PII replaced by placeholders. Integrated gradients attribute this distance to attention and feed-forward neurons, a scale-invariant specificity score isolates the PII-responsive ones, and zeroing them yields the pruned model.



 \item \textbf{Empirical evaluation.} Across five LLMs from three families and \underline{3} PII datasets, \method{} shows that pruning 1.6\% of neurons reduces PII inference accuracy by up to 7.4$\times$ and extraction recall by up to 13.9$\times$, outperforming random pruning. At a 3.2\% budget, \method{} matches data scrubbing with lower perplexity, while DEPN achieves lower leakage at the cost of at least 26$\times$ higher unseen-text perplexity.

\end{itemize}

\section{Related Work}
\label{sec:related}

\noindent{\bf\em PII Memorization and Extraction.}
LLMs can reproduce memorized training data verbatim, with leakage increasing with model size and data duplication and persisting after alignment \citep{carlini2021extracting,carlini2022quantifying,nasr2025scalable}. Fine-tuning exposes PII in proprietary records, with leakage depending on the updated parameters \citep{mireshghallah2022empirical,lukas2023analyzing}. Targeted attacks exploit auxiliary knowledge and repeated queries to extract additional PII \citep{kim2023propile,nakka2025pii}; scrubbed records can also serve as prompts for recovering masked values \citep{meng2025rr}. These findings show that private data remains extractable despite seemingly safe outputs.

\noindent{\bf\em Mitigating PII Leakage.}
\emph{Data scrubbing} masks or replaces identifiers before training \citep{yermilov2023privacy}, but missed entities remain exposed and masking may remove useful context \citep{lukas2023analyzing}. \emph{DP} clips per-example gradients and adds calibrated noise \citep{abadi2016deep,dwork2006differential}; it provides formal guarantees but may reduce utility and requires training-time intervention, including for LoRA fine-tuning \citep{yu2021differentially}. \emph{Post-training unlearning} uses shard retraining or optimization-based methods to remove selected data \citep{jang2023knowledge,zhang2024negative}. PII-specific methods include \textsc{Proactive Privacy Amnesia}, which replaces memorized PII with substitute memories \citep{kuo2025proactive}. However, forgotten information can resurface after subsequent model transformations \citep{hu2025unlearning,zhang2025catastrophic,lucki2024adversarial}.

\noindent{\bf\em Neuron Localization \& Privacy Editing.}
Transformer feed-forward layers store factual mappings \citep{geva2021transformer}, and integrated gradients help localize knowledge to neurons \citep{dai2022knowledge}. However, localized neurons may encode unrelated memories \citep{chang2024localization}. Selective pruning contrasts forget and retain sets \citep{wei2024assessing}. PII defenses include \textsc{DEPN} \citep{wu2023depn}, activation patching-based \textsc{APNEAP} \citep{wu2024mitigating}, and neuron-mask or polarity-aware editing \citep{zhengyuan2026ldede}. Unlike \method{}, these methods lack a PII-free reference of the same records.

\noindent{\bf\em Reference Models \& Public--Private Contrast.} LiRA detects memorization using shadow models trained with/without target samples \citep{carlini2022membership}, while task-vector editing removes learned behaviors in weight space \citep{ilharco2022editing}. Private prediction \citep{ginart2022submix} and prediction-difference clipping \citep{vinod2026invisibleink} use public--private contrasts at inference, adding runtime overhead unlike pruning neurons. Adapter-based access control routes queries between private and sanitized adapters \citep{shaaban2026securegate}, restricting access rather than removing memorized data.


\noindent{\bf\em Our Position.} Existing defenses sanitize data, modify training, or edit neurons without a PII-free reference. To our knowledge, \method{} is the first to use a JEPA-trained, PII-free reference for post-hoc neuron pruning without retraining.

\section{Preliminaries} \label{sec:problem}

\subsection{Background and Problem Setup} \label{sec:formulation}

\noindent{\bf\em PII Memorization in Fine-Tuned LLMs.}
Let $\mathcal{D}$ denotes an organization's text corpus. Each record $x=(w_1,\dots,w_T)$ contains PII entities $\mathcal{E}(x)$, where $(b,e,\tau)\in\mathcal{E}(x)$ identifies the span $w_{b:e}$ and entity type $\tau$ (e.g., \texttt{FIRSTNAME}). Fine-tuning a pretrained LLM $p_\theta$ on $\mathcal{D}$ minimizes the next-token cross-entropy: $\LM(x)=-\frac{1}{T}\sum_{t=1}^{T}\log p_\theta(w_t\mid w_{<t})$. The model may memorize these entities and reproduce their values verbatim, even without the surrounding context \citep{lukas2023analyzing}.

\vspace{1mm}
\noindent{\bf\em Parameter-Efficient Fine-Tuning.}
LoRA  freezes each pretrained weight matrix $W_0\in\mathbb{R}^{d_1\times d_2}$ and learns a low-rank update $\Delta W=BA$, where $B\in\mathbb{R}^{d_1\times r}$, $A\in\mathbb{R}^{r\times d_2}$, and $r\ll\min(d_1,d_2)$. The effective weight is $W_0+\Delta W$. Multiple adapters can therefore share the same frozen backbone while differing only in their learned updates.

\vspace{1mm}
\noindent{\bf\em Neurons and Privacy Editing.}
Each transformer layer $\ell$ has attention and feed-forward output projections, $W^\ell_{\mathrm{O}}$ and $W^\ell_{\mathrm{out}}$. Following the key--value memory view of feed-forward layers \citep{dai2022knowledge}, we define neuron $(\ell,j)$ as input coordinate $j$ of either projection, with activation $a^\ell_j$. Zeroing this activation at every token position is equivalent to zeroing the corresponding output-projection column (Appendix~\ref{app:background}). Privacy-neuron editing scores neurons by their contribution to PII-token likelihood and deactivates the highest-scoring ones. Without a PII-free reference, it cannot distinguish PII-specific neurons from those encoding context.

\vspace{1mm}
\noindent{\bf\em Joint-Embedding Predictive Objectives.}
LLM-JEPA \citep{huang2026llm} augments language modeling with a predictor $P_\phi$ that maps a context-view embedding toward a related target-view embedding using a cosine-based objective (Appendix~\ref{app:background}). The embedding $h(x)\in\mathbb{R}^{d}$ is the final-token hidden state. \method{} adapts this objective to train a public model on PII-free views, providing a reference for identifying and pruning PII-responsive neurons in the private model.

\subsection{Threat Model} \label{sec:threat}

We consider an organization deploying an LLM fine-tuned on proprietary records, with external adversaries seeking to extract or infer private information through its query interface.

\vspace{1mm}



\noindent{\bf\em Roles, Capabilities, and Goals.}
The \emph{data owner} has white-box access to the model, training records, and PII annotations. The \emph{external adversary} has black-box query access, can issue repeated queries, observe generated text and exposed token probabilities, and exploit auxiliary context and candidate values. Its goals are PII extraction \citep{carlini2021extracting}, attribute inference \citep{lukas2023analyzing}, and membership inference \citep{shokri2017membership}.
\begin{itemize}[leftmargin=*]
    \item \textbf{\em Extraction:} Search generated text for PII. Success requires recovering a training-record value not generated by the pretrained base model.
    
    \item \textbf{\em Attribute inference:} Mask a PII value in a partial record and rank candidates by record likelihood. Success means ranking the true value first.
    

    \item \textbf{\em Membership inference:} Determine whether a record was used for training.
\end{itemize}
The adversary has no access to model weights, the public reference, or private training records beyond auxiliary knowledge, and cannot modify the deployed model; white-box attacks are out of scope. Before deployment, the data owner uses inference attacks to identify leaked records for attribution.
\begin{figure*}[!t]
  \centering
\resizebox{0.9\linewidth}{!}{%
    \input{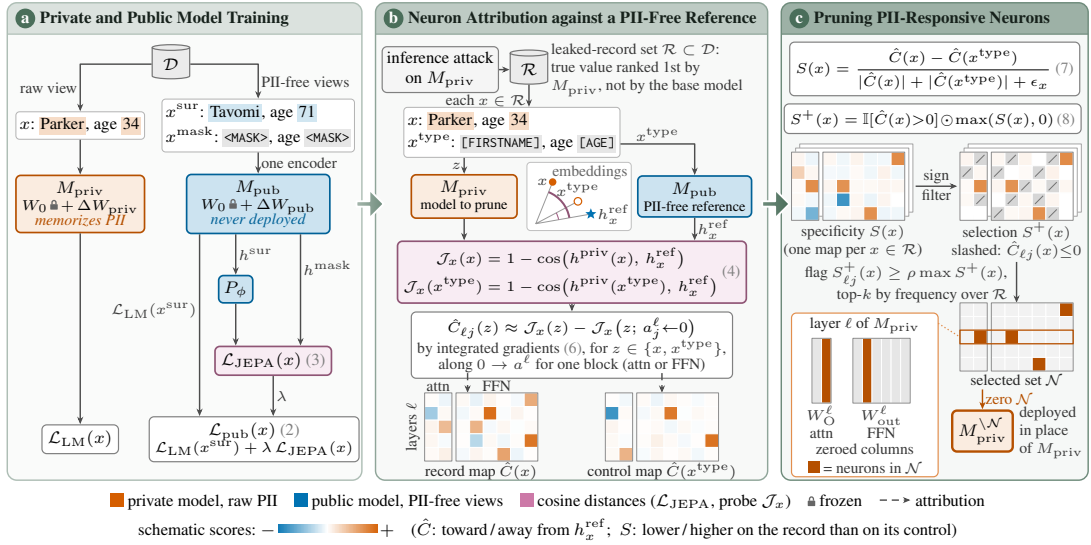}
}
\caption{Overview of \method{} workflow. \textbf{(a)}~Private $\Mpriv$ and public $\Mpub$ LoRA adapters share a frozen base model and train on raw records and PII-free views, respectively. \textbf{(b)}~For each leaked record, integrated gradients attribute the cosine distance between private-model embeddings of the record or its entity-type view and the public PII-free reference to attention and feed-forward neurons. \textbf{(c)}~PII-responsive neurons are zeroed, and the pruned model replaces $\Mpriv$ in deployment. Heatmaps and the embedding inset are schematic.}
  \label{fig:overview}
\end{figure*}

\section{Methodology} \label{sec:method}

\method{} mitigates PII leakage by pruning neurons in a fine-tuned LLM using a reference model trained on PII-free views of the same records. As shown in Fig.~\ref{fig:overview}, it comprises three stages: private/public model training, reference-guided neuron attribution, and selective pruning.

\subsection{Private and Public Model Training}
\label{sec:training}

We train two LoRA adapters on a shared frozen base model (Fig.~\ref{fig:overview}a): a {\sc private model} $\Mpriv$, fine-tuned on raw records, and a {\sc public model} $\Mpub$, trained on PII-free views and retained by the data owner as a reference. Given $\mathcal{D}$, we construct three views that differ only at annotated PII entities:
\begin{itemize}[leftmargin=*]
    \item \textbf{Raw:} $\mathcal{D}^{\mathrm{raw}}=\mathcal{D}$, preserving original entities.
    \item \textbf{Surrogate:} $\mathcal{D}^{\mathrm{sur}}$, replacing entities with synthetic values that preserve their surface form; $\xs$ denotes the surrogate view of record $x$.
    \item \textbf{Masked:} $\mathcal{D}^{\mathrm{mask}}$, replacing entities with \texttt{<MASK>}; $\xm$ denotes the masked view of $x$.
\end{itemize}
For each targeted matrix $W_0$, the adapters define as
\begin{equation}
\resizebox{6.9cm}{!}{$
\Delta W_{\mathrm{priv}}=B_{\mathrm{priv}}A_{\mathrm{priv}},
\quad
\Delta W_{\mathrm{pub}}=B_{\mathrm{pub}}A_{\mathrm{pub}},$}
\label{eq:adapters}
\end{equation}
yielding effective weights $W_0+\Delta W_{\mathrm{priv}}$ and $W_0+\Delta W_{\mathrm{pub}}$, respectively. Each adapter is optimized under its corresponding model. We denote the record embedding $h$ (Section~\ref{sec:formulation}) under these models by $h^{\mathrm{priv}}$ and $h^{\mathrm{pub}}$. Only the pruned private model is deployed; $\Mpub$ remains with the data owner.

\vspace{1mm}
\noindent{\bf\em Private-Model Fine-Tuning.}
We fine-tune $\Mpriv$ on $\mathcal{D}^{\mathrm{raw}}$ using the standard language-modeling loss $\LM$. Training on original records preserves domain information but can memorize PII \citep{lukas2023analyzing}. We leave this training procedure unchanged and mitigate leakage post hoc through neuron pruning (Section~\ref{sec:selection}).

\vspace{1mm}
\noindent{\bf\em Shape-Preserving Surrogate Generation.}
We synthesize surrogates rather than borrowing values from other records, which could raise real PII. Each digit is replaced with a uniformly random digit; alphabetic runs of at least three letters become pronounceable pseudo-words of similar length, alternating random consonants and vowels while preserving capitalization; shorter runs and punctuation remain unchanged (e.g., ``Parker''$\mapsto$``Tavomi'', ``4:56~PM''$\mapsto$``7:31~PM''). We redraw a surrogate if any word of at least three characters matches a corpus PII value, ignoring case. Views are constructed from $\mathcal{E}(x)$, obtained from raw annotations or a named-entity recognizer (NER). With NER-derived entities, missed PII remains unchanged; thus, PII removal is limited by NER recall.

\vspace{1mm}
\noindent{\bf\em JEPA Fine-Tuning on PII-Free Views.}
We train $\Mpub$ on surrogate text using language modeling and a joint-embedding predictive architecture (JEPA) objective \citep{lecun2022path,huang2026llm}. For each record, the surrogate view is the context and the masked view is the target:
\begin{align}
\mathcal{L}_{\mathrm{pub}}(x)
&=\LM(\xs)+\lambda\,\LJEPA(x),
\label{eq:pub}\\
\LJEPA(x)
&=1-\cos\bigl(P_{\phi}(h^{\mathrm{sur}}),
h^{\mathrm{mask}}\bigr),
\label{eq:jepa}
\end{align}
where $h^{\mathrm{sur}}=h^{\mathrm{pub}}(\xs)$, $h^{\mathrm{mask}}=h^{\mathrm{pub}}(\xm)$, and $P_{\phi}$ is a trainable affine predictor discarded after training. A single encoder embeds both views, with gradients flowing through both branches. Unlike standard LLM-JEPA, which uses context and target views in its predictive objective, our views are paired renderings of the same record, and the predictor is an affine map rather than a sequence of \texttt{[PRED]} tokens. Because the views differ only at entity spans, JEPA encourages representations that retain record context while reducing dependence on entity identity. Neither training term uses original PII values as model inputs, enabling $\Mpub$ to provide a PII-free reference, subject to entity-detection recall, for the subsequent attribution stage.

\subsection{Reference-Guided Neuron Attribution}
\label{sec:attribution}

Given the trained models (Section~\ref{sec:training}), we identify private-model neurons responsible for representation differences between leaked records and a PII-free reference (Fig.~\ref{fig:overview}b). Unlike likelihood-based privacy-neuron editing, our method attributes a representation-level discrepancy against an explicit PII-free reference.

\vspace{0.5mm}
\noindent{\bf\em Leaked-Record Selection.}
The data owner applies the inference attack (Section~\ref{sec:threat}) to $\Mpriv$ and selects records for which it ranks the true masked-entity value first, while the pretrained base model does not. This criterion isolates memorization-driven leaks. We select up to a fixed number of such records to form $\Aset\subset\mathcal{D}$, bounding attribution cost, which scales linearly with $|\Aset|$.

\vspace{0.5mm}
\noindent{\bf\em Cross-Model Probe Objective.}
For each $x\in\Aset$, let $\xc$ denote its entity-type view, replacing each entity $(b,e,\tau)\in\mathcal{E}(x)$ with a type placeholder (e.g., \texttt{[FIRSTNAME]}). The public model encodes $\xc$ into the PII-free reference $\htgt=h^{\mathrm{pub}}(\xc)$. We define the probe loss for the original and placeholder inputs as
\begin{equation}
  \begin{aligned}
    \Jx(x)&=1-\cos\bigl(h^{\mathrm{priv}}(x),\htgt\bigr),\\
    \Jx(\xc)&=1-\cos\bigl(h^{\mathrm{priv}}(\xc),\htgt\bigr).
  \end{aligned}
  \label{eq:probe}
\end{equation}
The record term measures the private model's deviation from the PII-free reference, while the control term captures PII-independent differences, such as syntax and domain vocabulary. The fixed reference requires only the private model to process probe inputs, introducing no trainable parameters.

\vspace{0.5mm}
\noindent{\bf\em Neuron Importance via Integrated Gradients.}
For $z\in\{x,\xc\}$, the contribution of neuron $(\ell,j)$ is
\begin{equation}
C_{\ell j}(z)=\Jx(z)-\Jx(z;\,a^\ell_j\!\leftarrow\!0),
\end{equation}
where $a^\ell_j$ is the neuron's activation (Section~\ref{sec:formulation}), zeroed at every token position. Positive contributions increase the discrepancy from the reference; negative contributions reduce it. Exact computation requires one forward pass per neuron, so following knowledge- and privacy-neuron localization \citep{dai2022knowledge,wu2023depn}, we approximate contributions using integrated gradients \citep{sundararajan2017axiomatic} along a path scaling the neuron's containing attention or feed-forward block from zero to its observed activations:
\begin{equation}
  \Chat_{\ell j}(z)=
  \sum_{t=1}^{|z|}a^\ell_{tj}\,
  \frac{1}{n}\sum_{i=0}^{n-1}
  \frac{\partial\Jx\bigl(z;\,\frac{i}{n}a^\ell\bigr)}
       {\partial a^\ell_{tj}},
  \label{eq:ig}
\end{equation}
where $|z|$ is the token count, $a^\ell=(a^\ell_{tj})_{t,j}$ denotes the block's activations, $\Jx(z;\alpha a^\ell)$ is the probe loss with those activations scaled by $\alpha$, and $n$ is the number of integration steps. By completeness, neuron contributions within a block approximately sum to the loss difference between its observed and zeroed activations. Applying Eq.~\eqref{eq:ig} to both inputs yields the record map $\Chat(x)$ and control map $\Chat(\xc)$. The latter captures PII-independent neuron contributions and supports the specificity analysis in the next stage. Unless indexed by $(\ell,j)$, scores denote neuron-wise importance maps.

\subsection{Pruning PII-Responsive Neurons}
\label{sec:selection}

We prune neurons that increase the private model's distance from the PII-free reference specifically on PII-bearing inputs (Fig.~\ref{fig:overview}c).

\vspace{0.5mm}
\noindent{\bf\em Scale-Invariant Specificity.}
Integrated-gradient magnitudes vary substantially across neurons, with a few high-activation units potentially dominating both contribution maps \citep{sun2024massive,yu2025superweightlargelanguage}. Thus, their difference alone cannot reliably identify PII-specific neurons. We normalize it by the total contribution:
\begin{equation}
S_{\ell j}(x)=
\frac{\Chat_{\ell j}(x)-\Chat_{\ell j}(\xc)}
{|\Chat_{\ell j}(x)|+|\Chat_{\ell j}(\xc)|+\epsilon_x},
\label{eq:ratio}
\end{equation}
where $\epsilon_x$ is the median total contribution across neurons for record $x$, stabilizing scores for negligible contributions. The resulting score lies in $(-1,1)$ and enables scale-invariant joint ranking of feed-forward and attention neurons.

\vspace{1mm}
\noindent{\bf\em Selecting PII-Responsive Neurons.}
We select neurons that increase the reference distance more for the original record than its entity-type view:
\begin{equation}
S^{+}_{\ell j}(x)=
\mathbb{I}\bigl[\Chat_{\ell j}(x)>0\bigr]
\max\bigl(S_{\ell j}(x),0\bigr).
\label{eq:select}
\end{equation}
The indicator excludes neurons that reduce record–reference distance; the positive part excludes neurons with control contribution matches or exceeds their record contribution. We call $S^{+}_{\ell j}(x)$ the \emph{selection score} and high-scoring neurons \emph{PII-responsive}. Following privacy-neuron editing \citep{wu2023depn}, we prune the largest positive contributors to the probe loss, preserving negative contributors that maintain PII-free behavior.


\vspace{1mm}
\noindent{\bf\em Cross-Record Aggregation and Pruning.}
For each $x\in\Aset$, we select neurons scoring at least $\rho\in(0,1]$ times the record-wise maximum: $S^{+}_{\ell j}(x)\geq\rho\max_{\ell',j'}S^{+}_{\ell'j'}(x)>0$. We select the $k$ neurons, across both attention and feed-forward blocks, flagged most frequently, breaking ties by the sum of flagged scores. Zeroing these neurons produces the pruned model $\Medit$, which replaces $\Mpriv$ at deployment. Pruning requires no retraining and modifies only $k$ columns of the private model's output projections, $W^{\ell}_{\mathrm{out}}$ and $W^{\ell}_{\mathrm{O}}$. The public reference remains unchanged and is never deployed.
 
\section{Performance Evaluation}
\label{sec:eval}

\noindent{\bf\em Datasets.}
We evaluate \method{} on three PII-containing datasets: ECHR \citep{chalkidis2019neural} (legal cases), Yelp Reviews \citep{zhang2015character} (user reviews), and ai4privacy \citep{ai4privacy_pii_masking_200k} (synthetic texts with annotated PII and entity types). We use 5,000 English training records per dataset.

\vspace{1mm}
\noindent{\bf\em LLMs.}
We evaluate five open-source LLMs across {\em 3} families: Llama-3.2-1B, Llama-3.2-3B, Llama-3.1-8B \citep{grattafiori2024llama}, Gemma-2-2B \citep{team2024gemma}, and Mistral-7B \citep{jiang2023mistral7b}. Implementation details appear in Appendix~\ref{app:settings}.

\vspace{1mm}
\noindent{\bf\em Attack Types.}
We evaluate extraction and inference attacks targeting person names \citep{carlini2021extracting,lukas2023analyzing}. Extraction samples model outputs unconditionally; inference ranks 50 candidate names for a masked name in each of 301 training records. We also evaluate LOSS-based and reference-based membership inference (Appendix~\ref{app:mia}); attack details appear in Appendix~\ref{app:attacks}.

\vspace{1mm}
\noindent{\bf\em Baselines.}
We compare \method{} against (i) the undefended private model $\Mpriv$; (ii) DEPN \citep{wu2023depn}, which prunes feed-forward neurons selected by PII-token likelihood attribution, often fewer than $k$; and (iii) random pruning, which removes $k$ neurons while matching \method{}'s per-layer allocation across attention and feed-forward blocks. This isolates the effect of targeted neuron selection on leakage.

\vspace{1mm}
\noindent{\bf\em  Metrics.}
Privacy metrics include extracted PII counts, extraction precision and recall, and inference top-1 accuracy. Utility is measured by perplexity (PPL) on seen training and unseen test records, over all and non-PII tokens, and the post-/pre-pruning PPL ratio. Formal definitions appear in Appendix~\ref{app:metrics}.


\subsection{Evaluating \method{} vs. Baselines}
\label{sec:main}

\begin{table*}[t]
  \centering\small\renewcommand{\arraystretch}{1.1}\setlength{\tabcolsep}{4pt}
  \begin{tabular}{@{}cl*{3}{rrr}@{}}
    \toprule
    & & \multicolumn{3}{c}{ai4privacy} & \multicolumn{3}{c}{ECHR} & \multicolumn{3}{c}{Yelp}\\
    \cmidrule(lr){3-5}\cmidrule(lr){6-8}\cmidrule(l){9-11}
    & & Inference & \multicolumn{2}{c}{Unseen PPL$\times$} & Inference & \multicolumn{2}{c}{Unseen PPL$\times$} & Inference & \multicolumn{2}{c}{Unseen PPL$\times$}\\
    \cmidrule(lr){4-5}\cmidrule(lr){7-8}\cmidrule(l){10-11}
    Model & Method & Acc.\,(\%) & All & Non-PII & Acc.\,(\%) & All & Non-PII & Acc.\,(\%) & All & Non-PII\\
    \midrule
    \multirow{8}{*}{\rotatebox[origin=c]{90}{Llama-3.2-1B}} & Unpruned & 11.96 & 1.000 & 1.000 & 26.58 & 1.000 & 1.000 & 35.22 & 1.000 & 1.000\\
    \cmidrule(l){2-11}
    & Random (0.8\%) & 10.30 & 1.039 & 1.030 & 24.92 & 1.073 & 1.075 & 31.89 & 1.048 & 1.047\\
    & Random (1.6\%) & 5.98 & 1.106 & 1.057 & 25.25 & 1.208 & 1.201 & 31.56 & 1.117 & 1.115\\
    & Random (3.2\%) & 3.65 & 1.477 & 1.316 & 21.26 & 1.176 & 1.164 & 25.91 & 1.328 & 1.328\\
    \cmidrule(l){2-11}
    & DEPN (0.13/0.11/0.03\%) & 2.66 & 1,942 & 3,970 & 1.99 & 2,033 & 1,781 & 2.66 & 755.9 & 750.0\\
    \cmidrule(l){2-11}
    & \method{} (0.8\%) & 5.98 & 1.148 & 1.037 & 23.59 & 1.131 & 1.105 & 3.99 & 55.343 & 54.761\\
    & \method{} (1.6\%) & 4.65 & 1.785 & 1.425 & 8.64 & 5.159 & 4.831 & 4.65 & 12.457 & 11.866\\
    & \method{} (3.2\%) & 2.99 & 5.442 & 3.289 & 4.65 & 27.382 & 27.078 & 1.66 & 46.942 & 44.700\\
    \midrule
    \multirow{8}{*}{\rotatebox[origin=c]{90}{Gemma-2-2B}} & Unpruned & 40.86 & 1.000 & 1.000 & 48.50 & 1.000 & 1.000 & 55.81 & 1.000 & 1.000\\
    \cmidrule(l){2-11}
    & Random (0.8\%) & 31.23 & 0.986 & 1.002 & 46.51 & 1.031 & 1.029 & 52.82 & 1.026 & 1.025\\
    & Random (1.6\%) & 21.59 & 1.152 & 1.208 & 40.20 & 1.062 & 1.055 & 49.83 & 1.066 & 1.063\\
    & Random (3.2\%) & 14.95 & 1.100 & 1.081 & 32.23 & 1.191 & 1.199 & 45.18 & 1.124 & 1.124\\
    \cmidrule(l){2-11}
    & DEPN (0.74/1.17/1.11\%) & 2.99 & 258.0 & 147.1 & 3.32 & 127.9 & 119.7 & 4.98 & 27.7 & 25.1\\
    \cmidrule(l){2-11}
    & \method{} (0.8\%) & 9.30 & 0.809 & 0.713 & 34.88 & 1.055 & 1.009 & 48.17 & 1.034 & 1.024\\
    & \method{} (1.6\%) & 8.64 & 0.855 & 0.745 & 26.25 & 1.155 & 1.083 & 43.85 & 1.108 & 1.092\\
    & \method{} (3.2\%) & 4.98 & 1.038 & 0.824 & 18.27 & 1.356 & 1.240 & 32.23 & 1.346 & 1.315\\
    \midrule
    \multirow{8}{*}{\rotatebox[origin=c]{90}{Llama-3.2-3B}} & Unpruned & 17.28 & 1.000 & 1.000 & 42.19 & 1.000 & 1.000 & 45.85 & 1.000 & 1.000\\
    \cmidrule(l){2-11}
    & Random (0.8\%) & 14.62 & 1.037 & 1.043 & 34.88 & 1.076 & 1.072 & 46.18 & 1.025 & 1.024\\
    & Random (1.6\%) & 9.63 & 1.175 & 1.149 & 35.88 & 1.097 & 1.087 & 43.52 & 1.061 & 1.059\\
    & Random (3.2\%) & 8.64 & 1.252 & 1.201 & 30.56 & 1.140 & 1.132 & 32.89 & 1.142 & 1.138\\
    \cmidrule(l){2-11}
    & DEPN (0.02/0.01/0.02\%) & 3.32 & 144.1 & 209.1 & 7.31 & 114.4 & 102.3 & 4.32 & 46.3 & 43.4\\
    \cmidrule(l){2-11}
    & \method{} (0.8\%) & 7.31 & 1.166 & 1.106 & 31.23 & 1.045 & 1.034 & 15.28 & 1.926 & 1.887\\
    & \method{} (1.6\%) & 5.32 & 1.451 & 1.326 & 24.92 & 1.158 & 1.125 & 10.96 & 3.090 & 2.987\\
    & \method{} (3.2\%) & 2.66 & 3.200 & 2.660 & 8.97 & 4.054 & 3.901 & 6.31 & 8.244 & 7.803\\
    \bottomrule
  \end{tabular}
  \caption{Inference accuracy and unseen perplexity of the pruned models.}
  \label{tab:main}
\end{table*}

\vspace{1mm}
\noindent{\bf\em  Evaluating Privacy and Utility.}
Table~\ref{tab:main} presents the inference accuracy and unseen PPL$\times$ of the pruned models. \method{} lowers the accuracy more than random pruning in every setting and at every budget, and also on Mistral-7B and Llama-3.1-8B (Appendix~\ref{app:models}). On Gemma-2-2B and ai4privacy at 0.8\%, for example, it cuts the accuracy from 40.86\% to 9.30\% while lowering the unseen PPL$\times$ to 0.809, whereas random pruning leaves 31.23\%. Because random pruning removes as many neurons from the same layers, this gap reflects which neurons \method{} selects, not how many. In contrast, DEPN reaches lower accuracies only by breaking the models, whose unseen PPL$\times$ is at least 27.7. Notably, the utility cost of \method{} depends on the model. On Gemma-2-2B, its unseen PPL$\times$ stays close to that of random pruning, whereas on the Llama-3.2 models it exceeds that of random pruning (at most 1.477) at 3.2\% and on Yelp, reaching 55.343, and on Llama-3.2-1B random pruning reaches similar accuracies at a similar perplexity on ai4privacy and ECHR. These findings indicate that \method{} locates the neurons that carry memorized names, but that on some models its utility cost limits how many of them can be pruned.

\begin{table}[t]
  \centering\small\setlength{\tabcolsep}{4pt}
  \begin{tabular}{@{}clrrr@{}}
    \toprule
    Dataset & Method & Extracted & \shortstack[r]{Precision\\(\%)} & \shortstack[r]{Recall\\(\%)}\\
    \midrule
    \multirow{4}{*}{\rotatebox[origin=c]{90}{ai4privacy}} & Unpruned & 474 & 8.04 & 24.01\\
    & Random & 489 & 5.07 & 24.77\\
    & DEPN (0.74\%) & 91 & 3.25 & 4.61\\
    & \method{} & 59 & 8.74 & 2.99\\
    \midrule
    \multirow{4}{*}{\rotatebox[origin=c]{90}{ECHR}} & Unpruned & 126 & 3.78 & 10.59\\
    & Random & 80 & 3.34 & 6.72\\
    & DEPN (1.17\%) & 8 & 0.26 & 0.67\\
    & \method{} & 9 & 2.82 & 0.76\\
    \bottomrule
  \end{tabular}
  \caption{Extraction attack against the pruned Gemma-2-2B at a budget of 1.6\%.}
  \label{tab:extraction}
\end{table}

\vspace{1mm}
\noindent{\bf\em  Evaluating Extraction.}
Table~\ref{tab:extraction} presents the extraction attack against Gemma-2-2B, the model that leaks most in Table~\ref{tab:main}, on ai4privacy and ECHR at a budget of 1.6\%. \method{} reduces the extracted names from 474 to 59 and from 126 to 9, whereas random pruning leaves 489 and 80. Recall falls accordingly, from 24.01\% to 2.99\% and from 10.59\% to 0.76\%, while precision changes little (8.74\% against 8.04\% and 2.82\% against 3.78\%), so the attacker recovers far fewer training names at a similar hit rate. In contrast, DEPN still extracts 91 and 8 names at an unseen PPL$\times$ of 258.0 and 127.9 (Table~\ref{tab:main}), against 0.855 and 1.155 for \method{}. Notably, \method{} already leaves only 68 and 20 names at 0.8\% (Appendix~\ref{app:full}). These findings indicate that the neurons selected with the inference attack carry the memorized names themselves rather than an attack-specific signal.

\begin{figure*}[!t]
  \centering
  \includegraphics[width=\textwidth]{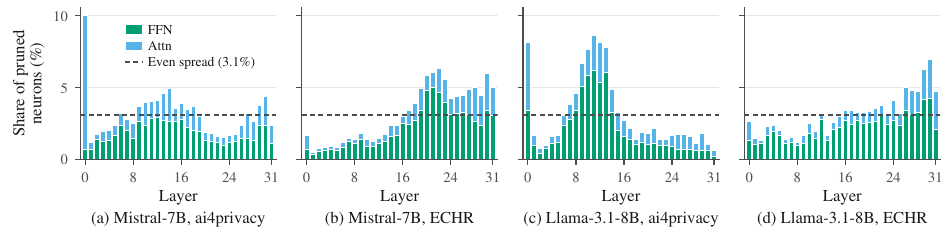}
  \caption{Distribution over layers of the neurons pruned by \method{} at a budget of 1.6\% ($k=9{,}437$ for both models); the bars of each panel sum to 100\%.}
  \label{fig:layer-dist}
\end{figure*}

\vspace{1mm}
\noindent{\bf\em  Evaluating Membership Inference.}
\method{} also weakens membership inference against Gemma-2-2B on ai4privacy (Appendix~\ref{app:mia}): across budgets of 0.8\%, 1.6\%, and 3.2\%, it lowers the AUC of the reference attack from 0.952 to 0.725--0.827, whereas random pruning only reaches 0.873--0.932, so the neurons that encode PII also carry much of the record-level memorization that membership inference exploits.

\vspace{1mm}
\noindent{\bf\em  Comparison with Public Models.}
Deploying a public model in place of the private model also limits leakage, but at a much higher unseen perplexity, whereas \method{} reaches a similar inference accuracy while keeping the perplexity close to that of the private model (Appendix~\ref{app:public}).

\vspace{-1.5mm}
\subsection{Ablation Studies}
\label{sec:ablation}

\begin{table}[t]
  \centering\small\renewcommand{\arraystretch}{1.3}\setlength{\tabcolsep}{1.7pt}
  \begin{tabular}{@{}clrrrrr@{}}
    \toprule
    & & Inference & \multicolumn{2}{c}{Unseen PPL$\times$} & \multicolumn{2}{c}{Seen PPL$\times$}\\
    \cmidrule(lr){4-5}\cmidrule(l){6-7}
    Model & Pruned & Acc.\,(\%) & All & Non-PII & All & Non-PII\\
    \midrule
    \multirow{4}{*}{\rotatebox[origin=c]{90}{Llama-3.2-1B}} & Unpruned & 11.96 & 1.000 & 1.000 & 1.000 & 1.000\\
    & FFN & 5.65 & 1.085 & 0.996 & 1.200 & 1.093\\
    & Attn & 2.99 & 1.701 & 1.534 & 2.298 & 2.134\\
    & Attn + FFN & 5.32 & 1.099 & 1.020 & 1.274 & 1.172\\
    \midrule
    \multirow{4}{*}{\rotatebox[origin=c]{90}{Mistral-7B}} & Unpruned & 49.17 & 1.000 & 1.000 & 1.000 & 1.000\\
    & FFN & 11.63 & 0.836 & 0.749 & 1.477 & 1.473\\
    & Attn & 16.94 & 0.768 & 0.711 & 1.426 & 1.527\\
    & Attn + FFN & 9.63 & 0.811 & 0.736 & 1.602 & 1.667\\
    \bottomrule
  \end{tabular}
  \caption{Effect of the pruned neuron type on ai4privacy ($k=1{,}000$ and $3{,}600$, 0.61\% of all attention and feed-forward neurons).}
  \label{tab:abl-type}
\end{table}

\noindent{\bf\em Evaluating Neuron Type Impact.}
Table~\ref{tab:abl-type} compares attention (Attn) and feed-forward (FFN) pruning at equal budgets. Attn-only pruning reduces Mistral-7B inference accuracy from 49.17\% to 16.94\%, but is less effective than FFN-only pruning on this model and reaches 2.99\% on Llama-3.2-1B only at a high unseen PPL$\times$ of 1.701. Joint ranking lets attribution allocate the budget across both types, assigning 37\% to Attn on Llama-3.2-1B and 51\% on Mistral-7B. It achieves the lowest accuracy on both models (5.32\% and 9.63\%), outperforming FFN-only pruning (5.65\% and 11.63\%) at similar unseen perplexity, except for Attn-only pruning on Llama-3.2-1B at substantial utility cost. All Mistral-7B variants reduce unseen PPL (PPL$\times$ 0.768--0.836) while increasing seen PPL (1.426--1.602), consistent with removing memorized training content. Thus, memorized PII spans both neuron types, with model-dependent contributions that joint ranking captures.

\vspace{1mm}
\noindent{\bf\em Evaluating Layer Distribution.}
Fig.~\ref{fig:layer-dist} shows the distribution of pruned Attn and FFN neurons across 32 layers of Mistral-7B and Llama-3.1-8B on ai4privacy and ECHR. Pruned neurons span all layers, but their distribution depends on the dataset. On ai4privacy, the first 16 layers contain 56.8\% and 73.6\% of pruned neurons, with peaks in layer 1 and layers 10--14. On ECHR, the last 16 layers contain 77.7\% and 67.0\%, with peaks in layers 20--23 for Mistral-7B and 26--30 for Llama-3.1-8B. Attn accounts for a larger share on ai4privacy (41.4\% and 34.6\%) than on ECHR (30.1\% and 28.1\%). Both model families exhibit the same dataset-dependent pattern, indicating that memorized PII has no fixed network location. Searching all layers and both neuron types avoids missing PII-related neurons outside predefined layers or FFN blocks.

\vspace{1mm}
\noindent{\bf\em Additional Ablations.}
Appendix~\ref{app:results} reports further ablations on Llama-3.2-1B and Gemma-2-2B. The first 0.08\% of neurons delivers most leakage reduction, while budgets above 0.61\% mainly harm utility (Appendix~\ref{app:abl-k}). Extraction-based or DEPN-style record selection performs similarly to inference-based selection, and most gains arise from a few records, with diminishing returns beyond about 50 (Appendices~\ref{app:abl-rec} and~\ref{app:abl-n}). Realistic surrogates increase public-model leakage without improving pruning (Appendix~\ref{app:abl-sur}). The JEPA reference matches the protection of a scrubbed-data reference at half the budget (Appendix~\ref{app:ref}), while entity-type views match masked views in leakage reduction with lower perplexity (Appendix~\ref{app:view}). Leakage reduction improves with LoRA rank and persists with 100 or 500 candidates (Appendices~\ref{app:abl-rank} and~\ref{app:cand}). Person-name-guided pruning has little effect on location and organization leakage (Appendix~\ref{app:types}).

\section{Conclusion}
\label{sec:conclusion}
We proposed \method, a post-hoc pruning framework that identifies neurons distinguishing leaked-record representations from those of a public model trained on PII-free views. Pruning these neurons reduces inference, extraction, and membership-inference attacks more effectively than random pruning while largely preserving utility without retraining. Unlike privacy-neuron editing and data scrubbing, \method achieves comparable privacy with much lower PPL costs.


\clearpage

\section{Limitations}
\label{sec:limitations}
\method{} has several limitations. Its utility cost depends on the model: the budget that preserves the perplexity of most LLMs substantially raises that of the smaller Llama models, so the budget must be chosen per model. Its PII-free views rely on entity annotations, and the entities that a named-entity recognizer misses reach the public model. Because the leaked-record set is built from an attack on person names, the neurons that \method{} prunes are specific to that entity type: pruning them leaves the leakage of other PII types largely unchanged, so protecting those types requires leaked records that contain them. Finally, pruned content may resurface after further fine-tuning or quantization.

\bibliography{custom}

\clearpage

\appendix

\section{Additional Results}
\label{app:results}

\subsection{Full Results}
\label{app:full}

Table~\ref{tab:main-full} extends Table~\ref{tab:main} with the seen PPL$\times$ on the training records, and Table~\ref{tab:extraction-full} extends Table~\ref{tab:extraction} to budgets of 0.8\% and 3.2\%. The extracted names fall with the budget, from 68 and 20 at 0.8\% to 29 and 5 at 3.2\% on ai4privacy and ECHR, whereas random pruning extracts 361--489 and 80--91 at every budget.

\begin{table*}[t]
  \centering\scriptsize\renewcommand{\arraystretch}{1.1}\setlength{\tabcolsep}{1.2pt}
  \begin{tabular}{@{}cl*{3}{rrrrr}@{}}
    \toprule
    & & \multicolumn{5}{c}{ai4privacy} & \multicolumn{5}{c}{ECHR} & \multicolumn{5}{c}{Yelp}\\
    \cmidrule(lr){3-7}\cmidrule(lr){8-12}\cmidrule(l){13-17}
    & & Inference & \multicolumn{2}{c}{Unseen PPL$\times$} & \multicolumn{2}{c}{Seen PPL$\times$} & Inference & \multicolumn{2}{c}{Unseen PPL$\times$} & \multicolumn{2}{c}{Seen PPL$\times$} & Inference & \multicolumn{2}{c}{Unseen PPL$\times$} & \multicolumn{2}{c}{Seen PPL$\times$}\\
    \cmidrule(lr){4-5}\cmidrule(lr){6-7}\cmidrule(lr){9-10}\cmidrule(lr){11-12}\cmidrule(lr){14-15}\cmidrule(l){16-17}
    Model & Method & Acc.\,(\%) & All & Non-PII & All & Non-PII & Acc.\,(\%) & All & Non-PII & All & Non-PII & Acc.\,(\%) & All & Non-PII & All & Non-PII\\
    \midrule
    \multirow{8}{*}{\rotatebox[origin=c]{90}{Llama-3.2-1B}} & Unpruned & 11.96 & 1.000 & 1.000 & 1.000 & 1.000 & 26.58 & 1.000 & 1.000 & 1.000 & 1.000 & 35.22 & 1.000 & 1.000 & 1.000 & 1.000\\
    \cmidrule(l){2-17}
    & Random (0.8\%) & 10.30 & 1.039 & 1.030 & 1.081 & 1.079 & 24.92 & 1.073 & 1.075 & 1.087 & 1.088 & 31.89 & 1.048 & 1.047 & 1.065 & 1.063\\
    & Random (1.6\%) & 5.98 & 1.106 & 1.057 & 1.197 & 1.147 & 25.25 & 1.208 & 1.201 & 1.246 & 1.234 & 31.56 & 1.117 & 1.115 & 1.183 & 1.177\\
    & Random (3.2\%) & 3.65 & 1.477 & 1.316 & 1.759 & 1.606 & 21.26 & 1.176 & 1.164 & 1.263 & 1.250 & 25.91 & 1.328 & 1.328 & 1.467 & 1.461\\
    \cmidrule(l){2-17}
    & DEPN (0.13/0.11/0.03\%) & 2.66 & 1,942 & 3,970 & 3,298 & 7,081 & 1.99 & 2,033 & 1,781 & 2,709 & 2,374 & 2.66 & 755.9 & 750.0 & 973.6 & 954.5\\
    \cmidrule(l){2-17}
    & \method{} (0.8\%) & 5.98 & 1.148 & 1.037 & 1.350 & 1.213 & 23.59 & 1.131 & 1.105 & 1.227 & 1.199 & 3.99 & 55.343 & 54.761 & 70.280 & 68.682\\
    & \method{} (1.6\%) & 4.65 & 1.785 & 1.425 & 2.388 & 1.955 & 8.64 & 5.159 & 4.831 & 6.724 & 6.260 & 4.65 & 12.457 & 11.866 & 15.982 & 15.051\\
    & \method{} (3.2\%) & 2.99 & 5.442 & 3.289 & 8.088 & 5.201 & 4.65 & 27.382 & 27.078 & 36.683 & 35.990 & 1.66 & 46.942 & 44.700 & 59.021 & 55.692\\
    \midrule
    \multirow{8}{*}{\rotatebox[origin=c]{90}{Gemma-2-2B}} & Unpruned & 40.86 & 1.000 & 1.000 & 1.000 & 1.000 & 48.50 & 1.000 & 1.000 & 1.000 & 1.000 & 55.81 & 1.000 & 1.000 & 1.000 & 1.000\\
    \cmidrule(l){2-17}
    & Random (0.8\%) & 31.23 & 0.986 & 1.002 & 1.186 & 1.206 & 46.51 & 1.031 & 1.029 & 1.064 & 1.067 & 52.82 & 1.026 & 1.025 & 1.078 & 1.074\\
    & Random (1.6\%) & 21.59 & 1.152 & 1.208 & 1.488 & 1.566 & 40.20 & 1.062 & 1.055 & 1.175 & 1.175 & 49.83 & 1.066 & 1.063 & 1.169 & 1.162\\
    & Random (3.2\%) & 14.95 & 1.100 & 1.081 & 1.781 & 1.861 & 32.23 & 1.191 & 1.199 & 1.446 & 1.475 & 45.18 & 1.124 & 1.124 & 1.337 & 1.325\\
    \cmidrule(l){2-17}
    & DEPN (0.74/1.17/1.11\%) & 2.99 & 258.0 & 147.1 & 697.9 & 493.9 & 3.32 & 127.9 & 119.7 & 263.7 & 260.1 & 4.98 & 27.7 & 25.1 & 45.7 & 40.4\\
    \cmidrule(l){2-17}
    & \method{} (0.8\%) & 9.30 & 0.809 & 0.713 & 1.640 & 1.577 & 34.88 & 1.055 & 1.009 & 1.179 & 1.138 & 48.17 & 1.034 & 1.024 & 1.174 & 1.156\\
    & \method{} (1.6\%) & 8.64 & 0.855 & 0.745 & 1.932 & 1.910 & 26.25 & 1.155 & 1.083 & 1.431 & 1.362 & 43.85 & 1.108 & 1.092 & 1.364 & 1.333\\
    & \method{} (3.2\%) & 4.98 & 1.038 & 0.824 & 2.622 & 2.488 & 18.27 & 1.356 & 1.240 & 1.952 & 1.837 & 32.23 & 1.346 & 1.315 & 1.851 & 1.782\\
    \midrule
    \multirow{8}{*}{\rotatebox[origin=c]{90}{Llama-3.2-3B}} & Unpruned & 17.28 & 1.000 & 1.000 & 1.000 & 1.000 & 42.19 & 1.000 & 1.000 & 1.000 & 1.000 & 45.85 & 1.000 & 1.000 & 1.000 & 1.000\\
    \cmidrule(l){2-17}
    & Random (0.8\%) & 14.62 & 1.037 & 1.043 & 1.096 & 1.100 & 34.88 & 1.076 & 1.072 & 1.163 & 1.158 & 46.18 & 1.025 & 1.024 & 1.054 & 1.052\\
    & Random (1.6\%) & 9.63 & 1.175 & 1.149 & 1.359 & 1.320 & 35.88 & 1.097 & 1.087 & 1.158 & 1.146 & 43.52 & 1.061 & 1.059 & 1.132 & 1.127\\
    & Random (3.2\%) & 8.64 & 1.252 & 1.201 & 1.527 & 1.476 & 30.56 & 1.140 & 1.132 & 1.267 & 1.258 & 32.89 & 1.142 & 1.138 & 1.289 & 1.278\\
    \cmidrule(l){2-17}
    & DEPN (0.02/0.01/0.02\%) & 3.32 & 144.1 & 209.1 & 315.3 & 492.3 & 7.31 & 114.4 & 102.3 & 188.1 & 166.3 & 4.32 & 46.3 & 43.4 & 68.3 & 62.6\\
    \cmidrule(l){2-17}
    & \method{} (0.8\%) & 7.31 & 1.166 & 1.106 & 1.670 & 1.637 & 31.23 & 1.045 & 1.034 & 1.164 & 1.149 & 15.28 & 1.926 & 1.887 & 2.545 & 2.462\\
    & \method{} (1.6\%) & 5.32 & 1.451 & 1.326 & 2.327 & 2.237 & 24.92 & 1.158 & 1.125 & 1.395 & 1.351 & 10.96 & 3.090 & 2.987 & 4.328 & 4.117\\
    & \method{} (3.2\%) & 2.66 & 3.200 & 2.660 & 5.867 & 5.279 & 8.97 & 4.054 & 3.901 & 6.033 & 5.792 & 6.31 & 8.244 & 7.803 & 11.992 & 11.145\\
    \bottomrule
  \end{tabular}
  \caption{Inference accuracy and unseen and seen perplexity of the pruned models at budgets of 0.8\%, 1.6\%, and 3.2\%.}
  \label{tab:main-full}
\end{table*}

\begin{table}[t]
  \centering\footnotesize\setlength{\tabcolsep}{2.6pt}
  \begin{tabular}{@{}clrrr@{}}
    \toprule
    Dataset & Method & Extracted & \shortstack[r]{Precision\\(\%)} & \shortstack[r]{Recall\\(\%)}\\
    \midrule
    \multirow{8}{*}{\rotatebox[origin=c]{90}{ai4privacy}} & Unpruned & 474 & 8.04 & 24.01\\
    & Random (0.8\%) & 471 & 7.89 & 23.86\\
    & Random (1.6\%) & 489 & 5.07 & 24.77\\
    & Random (3.2\%) & 361 & 10.48 & 18.29\\
    & DEPN (0.74\%) & 91 & 3.25 & 4.61\\
    & \method{} (0.8\%) & 68 & 9.43 & 3.44\\
    & \method{} (1.6\%) & 59 & 8.74 & 2.99\\
    & \method{} (3.2\%) & 29 & 8.03 & 1.47\\
    \midrule
    \multirow{8}{*}{\rotatebox[origin=c]{90}{ECHR}} & Unpruned & 126 & 3.78 & 10.59\\
    & Random (0.8\%) & 91 & 3.12 & 7.65\\
    & Random (1.6\%) & 80 & 3.34 & 6.72\\
    & Random (3.2\%) & 85 & 2.79 & 7.14\\
    & DEPN (1.17\%) & 8 & 0.26 & 0.67\\
    & \method{} (0.8\%) & 20 & 3.01 & 1.68\\
    & \method{} (1.6\%) & 9 & 2.82 & 0.76\\
    & \method{} (3.2\%) & 5 & 1.42 & 0.42\\
    \bottomrule
  \end{tabular}
  \caption{Extraction attack against the pruned Gemma-2-2B at budgets of 0.8\%, 1.6\%, and 3.2\%.}
  \label{tab:extraction-full}
\end{table}

\subsection{Additional Models}
\label{app:models}

\begin{table*}[t]
  \centering\small\renewcommand{\arraystretch}{1.3}\setlength{\tabcolsep}{4pt}
  \begin{tabular}{@{}clrrrrrrrrrr@{}}
    \toprule
    & & \multicolumn{5}{c}{ai4privacy} & \multicolumn{5}{c}{ECHR}\\
    \cmidrule(lr){3-7}\cmidrule(l){8-12}
    & & Inference & \multicolumn{2}{c}{Unseen PPL$\times$} & \multicolumn{2}{c}{Seen PPL$\times$} & Inference & \multicolumn{2}{c}{Unseen PPL$\times$} & \multicolumn{2}{c}{Seen PPL$\times$}\\
    \cmidrule(lr){4-5}\cmidrule(lr){6-7}\cmidrule(lr){9-10}\cmidrule(l){11-12}
    Model & Method & Acc.\,(\%) & All & Non-PII & All & Non-PII & Acc.\,(\%) & All & Non-PII & All & Non-PII\\
    \midrule
    \multirow{4}{*}{\rotatebox[origin=c]{90}{Mistral-7B}} & Unpruned & 49.17 & 1.000 & 1.000 & 1.000 & 1.000 & 59.47 & 1.000 & 1.000 & 1.000 & 1.000\\
    & Random & 29.57 & 1.022 & 1.019 & 1.261 & 1.314 & 53.49 & 1.027 & 1.026 & 1.103 & 1.105\\
    & DEPN & 1.99 & 145.1 & 413.6 & 453.4 & 1,889 & 5.98 & 26.0 & 28.3 & 52.5 & 59.1\\
    & \method{} & 6.64 & 0.888 & 0.797 & 2.030 & 2.181 & 33.89 & 0.985 & 0.963 & 1.336 & 1.323\\
    \midrule
    \multirow{4}{*}{\rotatebox[origin=c]{90}{Llama-3.1-8B}} & Unpruned & 12.29 & 1.000 & 1.000 & 1.000 & 1.000 & 36.21 & 1.000 & 1.000 & 1.000 & 1.000\\
    & Random & 10.30 & 1.099 & 1.079 & 1.140 & 1.117 & 36.21 & 1.043 & 1.038 & 1.074 & 1.068\\
    & DEPN & 2.66 & 1,474 & 2,392 & 2,424 & 3,954 & 4.32 & 2,560 & 2,349 & 3,543 & 3,225\\
    & \method{} & 5.98 & 1.147 & 1.072 & 1.533 & 1.404 & 28.24 & 1.110 & 1.086 & 1.241 & 1.207\\
    \bottomrule
  \end{tabular}
  \caption{Inference accuracy and perplexity of the pruned Mistral-7B and Llama-3.1-8B at a budget of 1.6\%.}
  \label{tab:models}
\end{table*}

Table~\ref{tab:models} presents the inference attack against Mistral-7B and Llama-3.1-8B on ai4privacy and ECHR at a budget of 1.6\%. \method{} lowers the accuracy of Mistral-7B from 49.17\% to 6.64\% on ai4privacy and from 59.47\% to 33.89\% on ECHR, and that of Llama-3.1-8B from 12.29\% to 5.98\% and from 36.21\% to 28.24\%. Random pruning lowers it far less, to 29.57\% and 53.49\% on Mistral-7B and to 10.30\% on Llama-3.1-8B and ai4privacy, and leaves Llama-3.1-8B on ECHR unchanged at 36.21\%. In contrast, DEPN lowers it further only by breaking both models, whose unseen PPL$\times$ is at least 26.0. Notably, \method{} lowers the unseen PPL$\times$ of Mistral-7B on ai4privacy to 0.888, unlike random pruning, while raising its seen PPL$\times$ to 2.030, consistent with removing memorized content rather than domain knowledge. On Llama-3.1-8B, it raises the unseen PPL$\times$ to 1.147 and 1.110, against 1.099 and 1.043 for random pruning, less than on the Llama-3.2 models at the same budget (Table~\ref{tab:main}). These findings indicate that \method{} carries over to larger models, where its utility cost is lower than on the smaller models of the same family.

\subsection{Membership Inference}
\label{app:mia}

\begin{table*}[t]
  \centering\small\setlength{\tabcolsep}{5pt}
  \begin{tabular}{@{}lrrrrrrr@{}}
    \toprule
    & & \multicolumn{2}{c}{LOSS attack} & \multicolumn{2}{c}{Reference attack} & \multicolumn{2}{c}{PPL$\times$}\\
    \cmidrule(lr){3-4}\cmidrule(lr){5-6}\cmidrule(l){7-8}
    Method & Budget & AUC & TPR@1\%FPR (\%) & AUC & TPR@1\%FPR (\%) & Unseen & Seen\\
    \midrule
    Unpruned & -- & 0.899 & 4.0 & 0.952 & 44.7 & 1.000 & 1.000\\
    \midrule
    Random & 0.8\% & 0.867 & 2.7 & 0.932 & 31.6 & 0.986 & 1.186\\
           & 1.6\% & 0.835 & 2.0 & 0.922 & 23.4 & 1.152 & 1.488\\
           & 3.2\% & 0.786 & 0.9 & 0.873 & 17.5 & 1.100 & 1.781\\
    \midrule
    DEPN & 0.74\% & 0.581 & 1.4 & 0.595 & 2.4 & 258.0 & 697.9\\
    \midrule
    DEPN ($\rho=0.04$) & 0.8\% & 0.557 & 1.4 & 0.569 & 2.1 & 1,413 & 4,092\\
              & 1.6\% & 0.545 & 0.7 & 0.561 & 1.6 & 2,575 & 7,799\\
              & 3.2\% & 0.540 & 0.7 & 0.552 & 1.8 & 1,986 & 6,252\\
    \midrule
    \method{} & 0.8\% & 0.737 & 1.2 & 0.827 & 11.0 & 0.809 & 1.640\\
              & 1.6\% & 0.693 & 1.1 & 0.784 & 9.5 & 0.855 & 1.932\\
              & 3.2\% & 0.649 & 1.3 & 0.725 & 5.7 & 1.038 & 2.622\\
    \bottomrule
  \end{tabular}
  \caption{Membership inference against the pruned Gemma-2-2B on ai4privacy.}
  \label{tab:mia}
\end{table*}

Table~\ref{tab:mia} presents membership inference against Gemma-2-2B on ai4privacy, comparing \method{} with random pruning and with DEPN, whose selection rule yields 2,181 feed-forward neurons at every budget unless its threshold ratio is lowered to $\rho=0.04$ to fill the budget. The unpruned private model exposes membership to both attacks, with an AUC of 0.899 under the LOSS attack and 0.952 under the reference attack. Across budgets of 0.8\%, 1.6\%, and 3.2\%, \method{} lowers these AUCs to 0.649--0.737 and 0.725--0.827, whereas random pruning only reaches 0.786--0.867 and 0.873--0.932. Since random pruning removes as many neurons from the same layers, this gap can only come from which neurons are removed: those that \method{} selects carry the membership signal. The reference attack, which calibrates each record's loss with the base model, is also far more confident: it identifies 44.7\% of the members at a 1\% FPR on the unpruned model, which \method{} reduces to 5.7--11.0\% against 17.5--31.6\% for random pruning, whereas the LOSS attack stays close to the 1\% chance level for every pruned model. At 0.8\% and 1.6\%, \method{} even lowers the unseen PPL$\times$ (0.809 and 0.855) while raising the seen PPL$\times$ (1.640 and 1.932), consistent with removing what the model memorized from its training records. In contrast, the lower AUCs of DEPN (0.540--0.595) stem from breaking the model, whose unseen PPL$\times$ reaches 258.0 to 2,575, whether or not it fills the budget. These findings indicate that the neurons that encode PII also carry much of the record-level memorization that membership inference exploits.

\subsection{Pruning Budget}
\label{app:abl-k}

\begin{table}[t]
  \centering\small\setlength{\tabcolsep}{3.5pt}
  \begin{tabular}{@{}rrrrrr@{}}
    \toprule
    & Inference & \multicolumn{2}{c}{Unseen PPL$\times$} & \multicolumn{2}{c}{Seen PPL$\times$}\\
    \cmidrule(lr){3-4}\cmidrule(l){5-6}
    $k$ & Acc.\,(\%) & All & Non-PII & All & Non-PII\\
    \midrule
    0 (unpruned) & 11.96 & 1.000 & 1.000 & 1.000 & 1.000\\
    125 & 6.98 & 1.007 & 0.975 & 1.059 & 1.009\\
    250 & 6.31 & 1.012 & 0.971 & 1.089 & 1.028\\
    500 & 5.65 & 1.024 & 0.969 & 1.141 & 1.069\\
    1,000 & 5.32 & 1.099 & 1.020 & 1.274 & 1.172\\
    2,000 & 5.32 & 1.292 & 1.126 & 1.574 & 1.374\\
    4,000 & 3.32 & 3.784 & 2.575 & 5.483 & 3.938\\
    \bottomrule
  \end{tabular}
  \caption{Effect of the pruning budget $k$ on Llama-3.2-1B and ai4privacy.}
  \label{tab:abl-k}
\end{table}

Table~\ref{tab:abl-k} presents the effect of the budget $k$ on Llama-3.2-1B. Pruning 125 to 1,000 neurons (0.08--0.61\% of 163,840) lowers the inference accuracy from 11.96\% to between 6.98\% and 5.32\%, with the largest drop from the first 125 neurons. Over this range, the unseen PPL$\times$ stays at or below 1.099 and its non-PII part within 0.969--1.020, whereas the seen PPL$\times$ rises to 1.274, consistent with pruning removing content specific to the training records. In contrast, larger budgets degrade utility sharply, to an unseen PPL$\times$ of 1.292 at $k=2{,}000$ and 3.784 at $k=4{,}000$, for little further reduction in accuracy (5.32\% and 3.32\%). Accuracy plateaus from $k=500$ on (5.65\% at $k=500$ against 5.32\% at $k=1{,}000$), indicating that a small fraction of neurons carries most of the leakage and that larger budgets mainly cost utility.

\subsection{Leaked-Record Selection}
\label{app:abl-rec}

\begin{table}[t]
  \centering\small\setlength{\tabcolsep}{1.1pt}
  \begin{tabular}{@{}lrrrrrr@{}}
    \toprule
    & & Inference & \multicolumn{2}{c}{Unseen PPL$\times$} & \multicolumn{2}{c}{Seen PPL$\times$}\\
    \cmidrule(lr){4-5}\cmidrule(l){6-7}
    Rule & Overlap & Acc.\,(\%) & All & Non-PII & All & Non-PII\\
    \midrule
    Unpruned & -- & 11.96 & 1.000 & 1.000 & 1.000 & 1.000\\
    Inference & 0 & 5.32 & 1.099 & 1.020 & 1.274 & 1.172\\
    Extraction & 71 & 4.98 & 1.064 & 1.007 & 1.260 & 1.184\\
    DEPN-style & 15 & 5.65 & 1.077 & 0.997 & 1.234 & 1.137\\
    \bottomrule
  \end{tabular}
  \caption{Effect of the leaked-record selection on Llama-3.2-1B and ai4privacy ($k=1{,}000$). Overlap: leaked records among the 301 evaluation records.}
  \label{tab:abl-rec}
\end{table}

Table~\ref{tab:abl-rec} replaces the inference-guided records with two alternatives (Appendix~\ref{app:selection}): the first 100 training records containing a name that the extraction attack recovers, and 100 records whose person names the private model ranks highly, sampled as in DEPN \citep{wu2023depn}. All three rules lower the accuracy from 11.96\% to 4.98--5.65\% at a similar perplexity cost (unseen PPL$\times$ 1.064--1.099); in this setting, pruning is thus insensitive to how the leaked records are found. We keep the inference rule for two reasons: its records are disjoint from the evaluation records by construction, whereas 71 extraction-guided and 15 DEPN-style records are evaluation records (Overlap), and its records are those that the private model demonstrably leaks. Attribution thus starts from verified leakage, and every reduction we report is measured on records that pruning never saw.

\subsection{Leaked-Record Count}
\label{app:abl-n}

\begin{table}[t]
  \centering\small\setlength{\tabcolsep}{3.5pt}
  \begin{tabular}{@{}rrrrrr@{}}
    \toprule
    & Inference & \multicolumn{2}{c}{Unseen PPL$\times$} & \multicolumn{2}{c}{Seen PPL$\times$}\\
    \cmidrule(lr){3-4}\cmidrule(l){5-6}
    $|\Aset|$ & Acc.\,(\%) & All & Non-PII & All & Non-PII\\
    \midrule
    Unpruned & 40.86 & 1.000 & 1.000 & 1.000 & 1.000\\
    \midrule
    \multicolumn{6}{@{}l}{\textit{1.6\% of the neurons ($k=4{,}686$)}}\\
    10 & 12.62 & 0.902 & 0.800 & 1.660 & 1.631\\
    25 & 14.62 & 0.853 & 0.783 & 1.557 & 1.558\\
    50 & 10.30 & 0.853 & 0.739 & 1.797 & 1.731\\
    100 & 8.64 & 0.855 & 0.745 & 1.932 & 1.910\\
    153 (all) & 7.31 & 0.870 & 0.733 & 2.042 & 1.969\\
    \midrule
    \multicolumn{6}{@{}l}{\textit{3.2\% of the neurons ($k=9{,}372$)}}\\
    10 & 9.30 & 0.910 & 0.784 & 1.984 & 1.975\\
    25 & 8.97 & 0.948 & 0.822 & 2.089 & 2.077\\
    50 & 6.64 & 0.967 & 0.793 & 2.329 & 2.249\\
    100 & 4.98 & 1.038 & 0.824 & 2.622 & 2.488\\
    153 (all) & 6.31 & 1.026 & 0.821 & 2.675 & 2.562\\
    \bottomrule
  \end{tabular}
  \caption{Effect of the number of leaked records $|\Aset|$ on Gemma-2-2B and ai4privacy.}
  \label{tab:abl-n}
\end{table}

Table~\ref{tab:abl-n} presents the effect of the number of leaked records on Gemma-2-2B at budgets of 1.6\% and 3.2\% of its neurons, using the first $|\Aset|$ records selected by the inference attack; the 153 records are all the leaked records that the attack finds. Even 10 records lower the accuracy from 40.86\% to 12.62\% and 9.30\%. Fewer than 50 records are nevertheless weaker than 100: at 3.2\%, 10 and 25 records leave 9.30\% and 8.97\% against 4.98\%, and at 1.6\%, 25 records leave 14.62\% against 8.64\%. From 50 records on, the accuracy comes close to that with 100 (10.30\% and 6.64\% against 8.64\% and 4.98\%), although the 50- and 100-record sets share only about half of their neurons. At 1.6\%, the unseen PPL$\times$ stays at 0.853--0.902 for every $|\Aset|$, whereas the seen PPL$\times$ rises from 1.557--1.660 with 10 or 25 records to 1.932--2.042 with 100 or more. These findings indicate that \method{} needs only a small sample of leaked records: a few already remove most of the leakage, and additional records bring diminishing returns.

\subsection{Surrogate Generator}
\label{app:abl-sur}

\begin{table}[t]
  \centering\small\setlength{\tabcolsep}{1.6pt}
  \begin{tabular}{@{}lrrrrrr@{}}
    \toprule
    & \multicolumn{2}{c}{\shortstack{Inference\\Acc.\,(\%)}} & \multicolumn{2}{c}{Unseen PPL$\times$} & \multicolumn{2}{c}{Seen PPL$\times$}\\
    \cmidrule(lr){2-3}\cmidrule(lr){4-5}\cmidrule(l){6-7}
    Generator & \shortstack{Public\\JEPA} & Pruned & All & Non-PII & All & Non-PII\\
    \midrule
    Pseudo-words & 1.33 & 5.32 & 1.099 & 1.020 & 1.274 & 1.172\\
    Faker & 2.99 & 6.31 & 1.105 & 1.013 & 1.269 & 1.155\\
    Hybrid & 4.32 & 6.31 & 1.096 & 1.017 & 1.263 & 1.161\\
    \bottomrule
  \end{tabular}
  \caption{Effect of the surrogate generator of the public model on Llama-3.2-1B and ai4privacy ($k=1{,}000$).}
  \label{tab:abl-sur}
\end{table}

The public model serves as a PII-free reference, so its surrogate view replaces every entity with a pseudo-word (Section~\ref{sec:training}). Table~\ref{tab:abl-sur} compares this generator with realistic values from the Faker library, used either for every entity type that Faker supports (Faker) or for all but name-like entities (Hybrid; Appendix~\ref{app:surrogates}). For each generator, it reports the inference accuracy of the public model itself, trained with the JEPA objective of Eq.~\eqref{eq:pub} (Public JEPA), and of the private model pruned with this public model as the reference (Pruned). Realistic surrogates make the public model itself leak more: its accuracy rises from 1.33\% with pseudo-words to 2.99\% with Faker and 4.32\% with Hybrid. In contrast, the generator barely affects pruning: the three references select largely the same neurons, and the pruned models reach 5.32--6.31\% at a similar perplexity cost. Realistic surrogates thus bring no benefit: they let the reference learn more of the training PII without improving pruning, whereas pseudo-words keep the reference closest to PII-free and prune at least as well.

\subsection{LoRA Rank}
\label{app:abl-rank}

\begin{table}[t]
  \centering\small\setlength{\tabcolsep}{3pt}
  \begin{tabular}{@{}rrrrrrr@{}}
    \toprule
    & \multicolumn{2}{c}{Inference Acc.\,(\%)} & \multicolumn{2}{c}{Unseen PPL$\times$} & \multicolumn{2}{c}{Seen PPL$\times$}\\
    \cmidrule(lr){2-3}\cmidrule(lr){4-5}\cmidrule(l){6-7}
    $r$ & Unpruned & Pruned & All & Non-PII & All & Non-PII\\
    \midrule
    4  & 6.64 & 4.98 & 1.107 & 1.033 & 1.190 & 1.099\\
    8  & 11.96 & 5.32 & 1.099 & 1.020 & 1.274 & 1.172\\
    16 & 17.94 & 7.31 & 1.039 & 0.960 & 1.337 & 1.228\\
    32 & 29.90 & 10.63 & 1.060 & 0.937 & 2.037 & 1.867\\
    \bottomrule
  \end{tabular}
  \caption{Effect of the LoRA rank $r$ on Llama-3.2-1B and ai4privacy ($k=1{,}000$).}
  \label{tab:abl-rank}
\end{table}

Table~\ref{tab:abl-rank} presents the effect of the rank $r$ of the private and public adapters. Higher ranks memorize more: the unpruned accuracy grows from 6.64\% at $r=4$ to 29.90\% at $r=32$. For $r\geq8$, pruning lowers the accuracy by 56--64\%, to 5.32--10.63\%, at an unseen PPL$\times$ of 1.039--1.099 and an unseen non-PII PPL$\times$ of at most 1.020, whereas the seen PPL$\times$ grows with the rank to 2.037 at $r=32$. At $r=4$, where the private model leaks least, the reduction is small (from 6.64\% to 4.98\%). Pruning thus becomes more effective as memorization grows: under the same budget, the share of leakage it removes rises from 25\% at $r=4$ to 64\% at $r=32$.

\subsection{Candidate Count}
\label{app:cand}

\begin{table}[t]
  \centering\small\setlength{\tabcolsep}{8pt}
  \begin{tabular}{@{}rrr@{}}
    \toprule
    & \multicolumn{2}{c}{Inference Acc.\,(\%)}\\
    \cmidrule(l){2-3}
    Candidates & Unpruned & Pruned\\
    \midrule
    50 & 11.96 & 5.32\\
    100 & 6.64 & 3.65\\
    500 & 1.99 & 0.66\\
    \bottomrule
  \end{tabular}
  \caption{Effect of the number of candidates on Llama-3.2-1B and ai4privacy ($k=1{,}000$).}
  \label{tab:cand}
\end{table}

Table~\ref{tab:cand} presents the inference attack against Llama-3.2-1B on ai4privacy with 50, 100, and 500 candidates per masked name. More candidates make the attack harder: the accuracy of the unpruned private model falls from 11.96\% to 6.64\% and 1.99\%. \method{} lowers it to 5.32\%, 3.65\%, and 0.66\%, a reduction of 56\%, 45\%, and 67\%. Although the accuracy itself falls sixfold as the candidate set grows tenfold, \method{} removes about half or more of the leakage at every count. These findings indicate that \method{} removes the memorized names themselves rather than an advantage tied to a particular candidate set, so its protection does not weaken as the attacker considers more candidates.

\subsection{Other Entity Types}
\label{app:types}

\begin{table}[t]
  \centering\small\setlength{\tabcolsep}{4pt}
  \begin{tabular}{@{}lrrrr@{}}
    \toprule
    & & \multicolumn{3}{c}{Inference Acc.\,(\%)}\\
    \cmidrule(l){3-5}
    Entity type & Records & Unpruned & DEPN & \method{}\\
    \midrule
    PERSON & 301 & 11.96 & 3.32 & 5.32\\
    LOC & 181 & 15.47 & 1.10 & 13.26\\
    ORG & 301 & 33.89 & 1.33 & 33.55\\
    PRODUCT & 251 & 47.01 & 0.00 & 39.44\\
    \bottomrule
  \end{tabular}
  \caption{Effect of pruning on the leakage of other entity types on Llama-3.2-1B and ai4privacy ($k=1{,}000$).}
  \label{tab:types}
\end{table}

Table~\ref{tab:types} presents the inference attack on three further entity types that Flair tags in ai4privacy, locations (LOC), organizations (ORG), and products (PRODUCT), against the pruned model of Appendix~\ref{app:abl-k}, whose neurons are located from person-name leaks. \method{} lowers the accuracy on person names from 11.96\% to 5.32\%, but leaves organizations unchanged (33.89\% and 33.55\%) and lowers locations only slightly, from 15.47\% to 13.26\%. Its 16\% reduction on products comes almost entirely from fragments of browser user-agent strings, such as ``Mozilla'' and ``Gecko'', that Flair tags as products. In contrast, DEPN lowers the accuracy of every type to at most 3.32\%, but only because it breaks the model, whose non-PII PPL$\times$ reaches 4,044. These findings indicate that the neurons \method{} prunes are specific to the entity type of the leaked records that guide it, so protecting other types requires leaked records of those types.

\subsection{JEPA versus Scrubbed Reference}
\label{app:ref}

\begin{table*}[t]
  \centering\small\setlength{\tabcolsep}{6pt}
  \begin{tabular}{@{}llrrrrr@{}}
    \toprule
    & & Inference & \multicolumn{2}{c}{Unseen PPL$\times$} & \multicolumn{2}{c}{Seen PPL$\times$}\\
    \cmidrule(lr){4-5}\cmidrule(l){6-7}
    $k$ & Reference & Acc.\,(\%) & All & Non-PII & All & Non-PII\\
    \midrule
    -- & Unpruned & 40.86 & 1.000 & 1.000 & 1.000 & 1.000\\
    \midrule
    \multirow{6}{*}{0.8\%} & Random & 31.23 & 0.986 & 1.002 & 1.186 & 1.206\\
     & Scrubbed & 14.29 & 0.757 & 0.684 & 1.432 & 1.399\\
    \cmidrule(l){2-7}
     & JEPA ($\lambda=0.5$) & 12.96 & 0.805 & 0.718 & 1.550 & 1.487\\
     & JEPA ($\lambda=1$) & 11.63 & 0.805 & 0.717 & 1.603 & 1.537\\
     & JEPA ($\lambda=2$, default) & 9.30 & 0.809 & 0.713 & 1.640 & 1.577\\
     & JEPA ($\lambda=4$) & 14.62 & 0.786 & 0.718 & 1.592 & 1.586\\
    \midrule
    \multirow{6}{*}{1.6\%} & Random & 21.59 & 1.152 & 1.208 & 1.488 & 1.566\\
     & Scrubbed & 10.30 & 0.764 & 0.685 & 1.623 & 1.621\\
    \cmidrule(l){2-7}
     & JEPA ($\lambda=0.5$) & 9.63 & 0.824 & 0.717 & 1.797 & 1.756\\
     & JEPA ($\lambda=1$) & 7.97 & 0.830 & 0.716 & 1.858 & 1.806\\
     & JEPA ($\lambda=2$, default) & 8.64 & 0.855 & 0.745 & 1.932 & 1.910\\
     & JEPA ($\lambda=4$) & 9.63 & 0.822 & 0.732 & 1.888 & 1.897\\
    \midrule
    \multirow{6}{*}{3.2\%} & Random & 14.95 & 1.100 & 1.081 & 1.781 & 1.861\\
     & Scrubbed & 8.64 & 0.823 & 0.718 & 1.955 & 1.977\\
    \cmidrule(l){2-7}
     & JEPA ($\lambda=0.5$) & 6.31 & 1.004 & 0.814 & 2.539 & 2.446\\
     & JEPA ($\lambda=1$) & 4.98 & 1.003 & 0.804 & 2.580 & 2.468\\
     & JEPA ($\lambda=2$, default) & 4.98 & 1.038 & 0.824 & 2.622 & 2.488\\
     & JEPA ($\lambda=4$) & 5.65 & 1.033 & 0.845 & 2.694 & 2.650\\
    \bottomrule
  \end{tabular}
  \caption{JEPA versus scrubbed PII-free reference on Gemma-2-2B and ai4privacy.}
  \label{tab:ref}
\end{table*}

Table~\ref{tab:ref} varies the public model that serves as the PII-free reference on Gemma-2-2B and ai4privacy: JEPA public models trained with $\lambda\in\{0.5,1,2,4\}$, and a public model trained on scrubbed records without the JEPA term. At every budget, the default JEPA reference ($\lambda=2$) leaves a lower inference accuracy than the scrubbed reference (9.30\%, 8.64\%, and 4.98\%, against 14.29\%, 10.30\%, and 8.64\%). The JEPA reference also removes more memorized content: it raises the seen PPL$\times$ more (2.622 against 1.955 at 3.2\%), while its non-PII PPL$\times$ stays below that of the unpruned model at every budget, so the rise of its all-token perplexity at 3.2\% comes from the PII tokens. In contrast, the scrubbed reference prunes more conservatively and needs twice the budget to match the JEPA reference: it reaches 8.64\% only at 3.2\%, the accuracy that the JEPA reference reaches at 1.6\%, at a similar unseen PPL$\times$ (0.823 and 0.855). Notably, $\lambda=1$ and $\lambda=2$ give the lowest accuracy at every budget, and all four JEPA references lower it far more than random pruning. These findings indicate that training the reference with the JEPA objective makes attribution more efficient, locating neurons whose removal erases more memorized PII, so the same protection requires a smaller edit.

\subsection{Entity-Type versus Masked View}
\label{app:view}

\begin{table}[t]
  \centering\small\setlength{\tabcolsep}{1.6pt}
  \begin{tabular}{@{}lrrrrr@{}}
    \toprule
    & Inference & \multicolumn{2}{c}{Unseen PPL$\times$} & \multicolumn{2}{c}{Seen PPL$\times$}\\
    \cmidrule(lr){3-4}\cmidrule(l){5-6}
    Reference / control & Acc.\,(\%) & All & Non-PII & All & Non-PII\\
    \midrule
    Unpruned & 11.96 & 1.000 & 1.000 & 1.000 & 1.000\\
    \midrule
    Masked / masked & 6.31 & 1.154 & 1.079 & 1.364 & 1.276\\
    Masked / type & 4.65 & 1.357 & 1.165 & 1.652 & 1.428\\
    Type / type & 5.32 & 1.099 & 1.020 & 1.274 & 1.172\\
    \bottomrule
  \end{tabular}
  \caption{Entity-type versus masked view on Llama-3.2-1B and ai4privacy ($k=1{,}000$).}
  \label{tab:view}
\end{table}

Table~\ref{tab:view} compares two PII-free views for computing the reference embedding and the control term: the entity-type view of \method{}, which replaces each PII value with a placeholder of its type, such as \texttt{[FIRSTNAME]}, and the masked view, which replaces every value with \texttt{<MASK>}. All three variants lower the accuracy from 11.96\% to a similar 4.65--6.31\%. In contrast, the entity-type view costs the least utility, raising the unseen PPL$\times$ to 1.099, against 1.154 when both use the masked view and 1.357 when only the reference embedding does. Notably, this mismatched variant is the most harmful, likely because pruning then also targets neurons that respond to the difference between the two views rather than to the PII. These findings indicate that pruning is most precise when the reference embedding and the control term use the same entity-type view, which removes the PII values but keeps their types, so that pruning targets the identities rather than the general knowledge of what kind of entity appears.

\subsection{Pruning versus Public Models}
\label{app:public}

\begin{table*}[t]
  \centering\small\setlength{\tabcolsep}{6pt}
  \begin{tabular}{@{}lrrrrrrr@{}}
    \toprule
    & Inference & \multicolumn{2}{c}{Unseen PPL$\times$} & \multicolumn{2}{c}{Seen PPL$\times$} & MIA & Extracted\\
    \cmidrule(lr){3-4}\cmidrule(lr){5-6}
    Model & Acc.\,(\%) & All & Non-PII & All & Non-PII & AUC & names\\
    \midrule
    Unpruned & 40.86 & 1.000 & 1.000 & 1.000 & 1.000 & 0.952 & 474\\
    \midrule
    Public, scrubbed & 5.32 & 1.617 & 1.294 & 2.125 & 1.511 & 0.861 & 56\\
    Public, JEPA ($\lambda=2$) & 3.65 & 2.200 & 0.928 & 4.274 & 1.613 & 0.772 & 0\\
    \midrule
    \method{} (0.8\%) & 9.30 & 0.809 & 0.713 & 1.640 & 1.577 & 0.827 & 68\\
    \method{} (1.6\%) & 8.64 & 0.855 & 0.745 & 1.932 & 1.910 & 0.784 & 59\\
    \method{} (3.2\%) & 4.98 & 1.038 & 0.824 & 2.622 & 2.488 & 0.725 & 29\\
    \bottomrule
  \end{tabular}
  \caption{Pruning the private model versus deploying a public model on Gemma-2-2B and ai4privacy.}
  \label{tab:public}
\end{table*}

Instead of pruning the private model, an organization could deploy a public model, which never learns the real PII values. Table~\ref{tab:public} compares the two options on Gemma-2-2B and ai4privacy, for the scrubbed public model and the JEPA public model. Both public models leak little (inference accuracies of 5.32\% and 3.65\%), but they raise the unseen PPL$\times$ to 1.617 and 2.200, and the scrubbed model also loses non-PII text (non-PII PPL$\times$ of 1.294). In contrast, \method{} at 3.2\% reaches a similar accuracy (4.98\%) with a lower perplexity than both public models, on all tokens (1.038) and on non-PII tokens (0.824). Notably, it also gives the lowest membership-inference AUC (0.725, against 0.861 and 0.772); only the JEPA public model, which never sees a real name, extracts fewer training names. These findings indicate that pruning offers a better privacy--utility trade-off than replacing the private model: it matches the privacy of a public model while retaining more of the utility gained from fine-tuning on the raw records.

\section{Detailed Experimental Settings}
\label{app:settings}

\subsection{Reproduction Hyperparameters}

\noindent{\bf\em Model Training.}
For each model and dataset, the private and public adapters are trained on the same records with identical optimization settings, isolating the effects of the model and dataset. All adapters use LoRA \citep{hu2021lora} with rank $r=8$, scaling factor $\alpha=2r$, and dropout $0.05$ on all attention and feed-forward projections of a 4-bit NF4-quantized base model \citep{dettmers2023qlora}. We train for 10 epochs using AdamW \citep{loshchilov2017decoupled}, an effective batch size of 32, and a learning rate of $5\times10^{-5}$, reduced to $2\times10^{-5}$ for Mistral-7B and Llama-3.1-8B, with linear decay. The base model uses double quantization and float16 computation. Training uses a per-device batch size of 8 and four gradient-accumulation steps. For Gemma-2-2B on ECHR, we instead use a batch size of 2 and 16 accumulation steps to accommodate its long documents and large vocabulary, preserving the effective batch size and number of optimization steps.

The public model uses the JEPA weight $\lambda=2$ and an affine predictor $P_{\phi}$; we also evaluate $\lambda\in\{0.5,1,4\}$. The scrubbed public model uses only the language-modeling loss, with Flair-recognized entities replaced by \texttt{<MASK>}. For ai4privacy, entity sets $\mathcal{E}(x)$ come from gold annotations; for ECHR and Yelp, we use the Flair named-entity recognizer \citep{akbik2019flair}, following \citet{lukas2023analyzing}. Each surrogate is redrawn at most 50 times.

\vspace{1mm}
\noindent{\bf\em Attribution and Record Selection.}
We attribute leakage using integrated gradients with $n=20$ steps, following DEPN \citep{wu2023depn}. Adapters are loaded on their 4-bit base models with unmerged LoRA weights, as during training. Inputs are truncated to 256 tokens, and attribution is computed at all token positions. The stabilizer $\epsilon_x$ is set to the per-record median of the total contribution.

For each setting, $\Aset$ contains the first 100 training records, in training order, outside the 301-record inference evaluation set, for which $\Mpriv$ ranks the masked name first in the inference attack but the pretrained base model does not. If fewer than 100 records meet this criterion, we use all qualifying records. Thus, attribution records are disjoint from evaluation records.

\vspace{1mm}
\noindent{\bf\em Pruning.}
We set the selection threshold to $\rho=0.1$, following the released DEPN implementation \citep{wu2023depn}. We evaluate pruning budgets of $0.8\%$, $1.6\%$, and $3.2\%$ of each model's combined attention and feed-forward neurons; most ablations use $0.61\%$. For Llama-3.2-1B, which has 163,840 such neurons, these budgets correspond to 1,311, 2,621, and 5,243 neurons, respectively, while $0.61\%$ corresponds to 1,000 neurons. Random pruning uses seed 42. All remaining settings follow \citet{lukas2023analyzing}.

\subsection{DEPN Adaptation}
\label{app:depn}
DEPN \citep{wu2023depn} was proposed for BERT, where it attributes the probability of a PII token at a masked position to the feed-forward neurons. We port it to decoder-only LLMs with three changes. Since a causal LLM has no masked position, we attribute the negative log-likelihood of the PII tokens given the text that precedes them. We scale the activations at all positions rather than at a single target position, because pruning zeroes a neuron at every position. Instead of its own selection of memorized names, DEPN uses the same leaked records as \method{}, so the two methods differ only in how they attribute and select neurons. Everything else follows the original method: DEPN attributes feed-forward neurons only, with integrated gradients over $n=20$ steps, keeps a neuron when it reaches a fraction $\rho=0.1$ of a record's largest attribution, selects the neurons kept in the most records, and prunes them by zeroing. Because this rule keeps few neurons on our models, DEPN often selects fewer than $k$ neurons; we then prune all of them. For membership inference (Appendix~\ref{app:mia}), we also evaluate DEPN with $\rho$ lowered to $0.04$, the largest value that fills the 3.2\% budget, and keep the top $k$ neurons of its ranking.

\subsection{Attack Settings}
\label{app:attacks}
The extraction attack draws 10,240 unconditional samples of up to 256 tokens with top-40 sampling from the attacked model, tags them with Flair, and counts a PERSON mention as extracted if it occurs in the training records and not in samples of the pretrained base model. The inference attack follows \citet{lukas2023analyzing} on 301 training records that contain a PERSON name: one name is masked, the remaining names are filled in by a masked language model, and the attacker ranks 50 candidates, the true name and 49 names sampled from the training records, by the model's perplexity on the completed record. For other entity types (Appendix~\ref{app:types}), the attack masks one Flair LOC, ORG, or PRODUCT mention per record and draws the 49 other candidates from the training mentions of the same type, on up to 300 training records outside the leaked records; the non-PII PPL$\times$ of this analysis is computed on 300 test records. For the candidate-count analysis, the attacker ranks 100 or 500 candidates instead, with the additional names also sampled from the training records. Membership inference uses the first 1,000 training records as members and 1,000 test records as non-members; the LOSS attack scores a record by its negative mean token loss under the model, and the reference attack by the loss under the pretrained base model minus the loss under the model.

\subsection{Surrogate Generators}
\label{app:surrogates}
The surrogate-generator ablation (Appendix~\ref{app:abl-sur}) compares three ways of filling the entity spans of the surrogate view; the masked view and all other training settings are unchanged.
\begin{itemize}
  \item \textbf{Pseudo-words} (the default of \method{}): every entity value is rewritten character by character, as described in Section~\ref{sec:training}.
  \item \textbf{Faker}: every entity type with a matching generator in the Faker library (version 40.38.0, locale \texttt{en\_US}; 44 entity types of ai4privacy) receives a realistic value of that type, e.g., a first name for \texttt{FIRSTNAME}, an e-mail address for \texttt{EMAIL}, and a date for \texttt{DATE}; the remaining types, such as \texttt{GENDER}, fall back to pseudo-words.
  \item \textbf{Hybrid}: name-like entities (\texttt{FIRSTNAME}, \texttt{MIDDLENAME}, \texttt{LASTNAME}, \texttt{USERNAME}, \texttt{STREET}, \texttt{CITY}, and \texttt{COMPANYNAME}) receive pseudo-words, e-mail addresses receive a pseudo-word user name with a Faker mail domain, and all other types are filled as in Faker.
\end{itemize}
All three generators share the collision filter of Section~\ref{sec:training}: a value is redrawn, up to 50 times, while one of its words of three or more characters occurs, ignoring case, among the PII values of the training data. Each record uses its own seeded generator, and repeated mentions of the same value within a record receive the same surrogate. For example, ``Parker Friesen'' becomes ``Luleh Adefagam'' with pseudo-words, ``Brittney Reeves'' with Faker, and ``Avurem Karecupu'' with the hybrid, and ``4:56~PM'' becomes ``7:66~PM'', ``08:18:09'', and ``04:04:48'', respectively: pseudo-words preserve the format but not the validity of a value, whereas Faker produces valid values in its own format. After 50 redraws, 1,877 Faker values and 1,845 hybrid values still share an alphabetic word with the training PII, mostly common words such as e-mail domains and state names, whereas no pseudo-word does.

\subsection{Leaked-Record Selection Rules}
\label{app:selection}
The leaked-record ablation (Appendix~\ref{app:abl-rec}) compares three ways of choosing the 100 training records whose attributions guide pruning; attribution, ranking, and the budget are unchanged.
\begin{itemize}
  \item \textbf{Inference attack} (the default of \method{}): the data owner runs the inference attack of Section~\ref{sec:threat} against the private model on the training records outside the 301 evaluation records. In each record, one person name is masked, and the record counts as leaked when the private model ranks the true name first among the 50 candidates while the pretrained base model does not. The first 100 leaked records, in training order, form $\Aset$.
  \item \textbf{Extraction attack}: the data owner runs the extraction attack (Appendix~\ref{app:attacks}) against the private model, which yields the person names that the private model generates and the base model does not. The first 100 training records, in training order, that contain one of these names form $\Aset$; the evaluation records are not excluded.
  \item \textbf{DEPN-style}: following the data selection of DEPN \citep{wu2023depn}, person names are detected in the training text with the named-entity chunker of NLTK \citep{bird2009natural}, and a name counts as memorized when the mean reciprocal rank of its tokens under the private model, given the preceding 20 words, exceeds 0.4. Records that contain a memorized name are sampled at random until 100 distinct records are selected. Unlike the other two rules, this test involves no base model, so frequent words that the chunker tags as names also count as memorized; for example, 29 of the 101 sampled names are the greeting ``Dear''.
\end{itemize}
The inference attack is the only rule whose records are disjoint from the evaluation records by construction: 71 of the extraction-guided records and 15 of the DEPN-style records are evaluation records.

\subsection{Evaluation Metrics}
\label{app:metrics}
We use the following metrics to assess privacy and utility:
\begin{itemize}
  \item \textbf{Extracted PII:} the number of distinct PII values that the extraction attack recovers from the training records.
  \item \textbf{Precision:} the fraction of the PII values extracted from the generated samples that occur in the training records, indicating actual leakage.
  \item \textbf{Recall:} the fraction of the PII values in the training records that the attack recovers, reflecting the extent of the leakage.
  \item \textbf{Inference accuracy:} the fraction of attacked records for which the true name is the lowest-perplexity candidate among the 50.
  \item \textbf{Membership AUC and TPR@1\%FPR:} the area under the ROC curve of a membership-inference attack (0.5 is chance) and the fraction of members it identifies while flagging 1\% of the non-members.
  \item \textbf{Perplexity (PPL):} the exponentiated mean negative log-likelihood over a set $\mathcal{T}$ of evaluated tokens; lower values indicate better predictive performance:
  \begin{equation*}
    \mathrm{PPL}=\exp\Bigl(-\frac{1}{|\mathcal{T}|}\sum_{(x,t)\in\mathcal{T}}\log p(w_t\mid w_{<t})\Bigr),
  \end{equation*}
  where $(x,t)$ indexes token $w_t$ of a held-out record $x$. For non-PII PPL, $\mathcal{T}$ contains only the tokens outside the entities of $\mathcal{E}(x)$. We report the ratio PPL$\times$ of the PPL after pruning to the PPL before pruning. Perplexity is computed on the first 4,000 test records of each dataset, which are never used for training, truncated to 1,024 tokens, and, in the ablations, also on the 5,000 training records (seen perplexity); the half-widths of the 95\% bootstrap confidence intervals over records are below 5\% of the reported non-PII PPL.
\end{itemize}

\subsection{Computing Environment}
All experiments ran on a SLURM-managed cluster with NVIDIA H100 PCIe GPUs (80\,GB). Extraction samples are generated deterministically with a fixed seed on the same GPU type, so the compared models differ only in their weights.

\subsection{Implementation Details}
The framework builds on the implementation of \citet{lukas2023analyzing} and Hugging Face Transformers \citep{wolf2020transformers}; named-entity recognition uses Flair \citep{akbik2019flair} with the \texttt{ontonotes-large} model and the settings of \citet{lukas2023analyzing}. All attacks and perplexity evaluations load each adapter as it was trained, on the 4-bit NF4 base model with unmerged LoRA weights; the base-model references use the same quantized base model without LoRA. Neurons are accessed through forward hooks on the input of each layer's output projections $W^{\ell}_{\mathrm{out}}$ and $W^{\ell}_{\mathrm{O}}$. Pruning a neuron zeroes this input at every position, which is equivalent to zeroing the corresponding column of the merged projection $W_0+\Delta W_{\mathrm{priv}}$ in $\Mpriv$.

\section{Background Details}
\label{app:background}

\subsection{Feed-Forward and Attention Neurons}
With hidden size $d$, the feed-forward block of layer $\ell$ maps its input $v\in\mathbb{R}^{d}$ at each position to
\begin{equation*}
  \mathrm{FFN}^{\ell}(v)=W^{\ell}_{\mathrm{out}}\,a^{\ell},\qquad a^{\ell}=f\bigl(W^{\ell}_{\mathrm{in}}v\bigr)\in\mathbb{R}^{d_{\mathrm{ff}}},
\end{equation*}
where $W^{\ell}_{\mathrm{in}}\in\mathbb{R}^{d_{\mathrm{ff}}\times d}$ and $W^{\ell}_{\mathrm{out}}\in\mathbb{R}^{d\times d_{\mathrm{ff}}}$ are the input and output projections and $f$ is an elementwise nonlinearity; each coordinate $j$ of $a^{\ell}$ is a feed-forward neuron, and zeroing it at every position is equivalent to zeroing the $j$-th column of $W^{\ell}_{\mathrm{out}}$. Gated variants \citep{shazeer2020glu} add a second input projection $W^{\ell}_{\mathrm{gate}}$ and compute $a^{\ell}=f\bigl(W^{\ell}_{\mathrm{gate}}v\bigr)\odot\bigl(W^{\ell}_{\mathrm{in}}v\bigr)$, where $\odot$ is the elementwise product; only $a^{\ell}$ changes, so neurons and their removal are defined identically. In the multi-head self-attention block, $H$ heads of dimension $d_{\mathrm{h}}$ produce outputs that are concatenated into $o^{\ell}\in\mathbb{R}^{Hd_{\mathrm{h}}}$ at each position and mapped back to $\mathbb{R}^{d}$ by the output projection $W^{\ell}_{\mathrm{O}}\in\mathbb{R}^{d\times Hd_{\mathrm{h}}}$; each coordinate of $o^{\ell}$ is an attention neuron, and zeroing it at every position is equivalent to zeroing the corresponding column of $W^{\ell}_{\mathrm{O}}$. The attention and feed-forward neurons of a layer are indexed jointly, so $a^{\ell}_{j}$ denotes the activation of a neuron of either kind.

\subsection{LLM-JEPA Objective}
LLM-JEPA \citep{huang2026llm} complements the language-modeling loss with an embedding-space term that predicts the representation of one view of the data (the target) from that of a related view (the context). With $h^{(1)}$ and $h^{(2)}$ the embeddings of the context view $x^{(1)}$ and the target view $x^{(2)}$, the objective is
\begin{equation}
  \begin{aligned}
    &\mathcal{L}_{\mathrm{LLM\text{-}JEPA}}=\LM(x^{(1)})\\
    &\qquad+\lambda\bigl[1-\cos\bigl(P_{\phi}(h^{(1)}),\,h^{(2)}\bigr)\bigr],
  \end{aligned}
  \label{eq:llmjepa}
\end{equation}
where $P_{\phi}$ is a predictor with parameters $\phi$ and $\lambda$ balances the two terms. In the original formulation, the two views are, for example, a natural-language description and the corresponding code.

%
%

\end{document}